\documentclass[manuscript, table, nonacm]{acmart}

\usepackage{xcolor}
\usepackage{amsmath}
\usepackage{graphicx}
\usepackage{caption}
\usepackage{subcaption}
\usepackage{verbatim}
\usepackage{latexsym}
\usepackage{setspace}
\usepackage{mathtools}
\usepackage{bbm}

\newtheorem{definition}{Definition}
\newtheorem{system}{System}
\newtheorem{theorem}{Theorem}
\newtheorem{proposition}{Proposition}
\newtheorem{problem*}{Problem}
\newtheorem{lemma}{Lemma}
\newtheorem{corollary}{Corollary}

\newtheorem{remark}{Remark}
\newtheorem{assumption}{Assumption}

\begin{document}
\title{Mean Field Behaviour of Collaborative Multi-Agent Foragers
}

\author{Daniel Jarne Ornia}
\email{d.jarneornia@tudelft.nl}
\affiliation{%
  \institution{DCSC, Delft University of Technology}
  \streetaddress{Mekelweg 2}
  \city{Delft}
  \state{The Netherlands}
  \postcode{2628CD}
}
\author{Pedro J Zufiria}
\email{pedro.zufiria@upm.es}
\affiliation{%
  \institution{Universidad Politecnica de Madrid}
  \streetaddress{Ciudad Universitaria s/n}
  \city{Madrid}
  \state{Spain}
  \postcode{28040}
}
\author{Manuel Mazo Jr.}
\email{m.mazo@tudelft.nl}
\affiliation{%
  \institution{DCSC, Delft University of Technology}
  \streetaddress{Mekelweg 2}
  \city{Delft}
  \state{The Netherlands}
  \postcode{2628CD}
}

\begin{abstract}
Collaborative multi-agent robotic systems where agents coordinate by modifying a shared environment often result in undesired dynamical couplings that complicate the analysis and experiments when solving a specific problem or task. Simultaneously, biologically-inspired robotics rely on simplifying agents and increasing their number to obtain more efficient solutions to such problems, drawing similarities with natural processes. In this work we focus on the problem of a biologically-inspired multi-agent system solving collaborative foraging. We show how mean field techniques can be used to re-formulate such a stochastic multi-agent problem into a deterministic autonomous system. This de-couples agent dynamics, enabling the computation of limit behaviours and the analysis of optimality guarantees. Furthermore, we analyse how having finite number of agents affects the performance when compared to the mean field limit and we discuss the implications of such limit approximations in this multi-agent system, which have impact on more general collaborative stochastic problems.
\end{abstract}
\maketitle
\thispagestyle{empty}
\section{INTRODUCTION}

Smaller processors and faster communications are pushing towards larger multi-agent systems with simple agents for solving large complex problems in a decentralised fashion. Be it large groups of autonomous cars driving in an urban setting or groups of nano-agents used in biomedical applications, there is a drive to increase the amount of collaborative agents in such settings, for either necessity (as in the case of vehicle traffic) or efficiency (in problems where more agents translate in better solutions). 
In the past two decades there has been growing interest in biologically inspired methods for solving decentralised coordination problems for large groups of simple agents. Inspiration is often drawn from the behaviour of ants, birds, bees or fish, for example \cite{beni1993swarm,dorigo2007swarm,blum2008swarm,kennedy2006swarm}. These biological systems seem to have developed inherent robustness towards problems such as individual agent errors, malfunction or communication disruptions. Furthermore, some of them have evolved to be extremely resource-efficient, that being in time, energy or information transmission.

We focus our attention in this paper on a particular subclass of bio-inspired multi-agent stochastic coordination problems: foraging. {Foraging is the problem of locating an unknown target, e.g. a source of food, in an unknown environment, and exploiting the shortest path to such target from a given initial location, e.g. a nest, with the goal of depleting the food source as fast as possible}. For an extensive set of stochastic multi-agent methods the foraging problem has served as both a benchmark but also a study subject on itself, given the combined nature of exploration plus optimization that the problem presents\cite{zedadra2017multi}. Naturally, many of the biological systems capable of solving foraging problems present some (degree of) de-centralised behaviour. Ants, for example, communicate with each-other only by depositing pheromones on the environment, and achieve global coordination by combining the individual contributions of all members of the swarm without the need of centralised instructions.  The mechanism of communicating indirectly through enviromental marking is known as \emph{stigmergy}.

Ants and bees make use of stigmergy methods to coordinate with other members of the swarm~\cite{grasse1959reconstruction,bernasconi1999cooperation} to solve specific tasks, for example foraging for food~\cite{carroll1973ecology,traniello1989foraging}. This kind of cooperative behaviour has been modelled for social insects such as ants \cite{watmough1995modelling,schweitzer1997active,meyer2008tale}, but also bees\cite{de1998modelling,sumpter2003modelling} in more general frameworks (see as well the work of Resnick \cite{resnick1997turtles} for an extensive analysis and application of such behaviours).

Ant-inspired heuristics have been widely used to solve foraging problems in a distributed fashion, sparking a whole branch of stochastic optimization algorithms: Ant Colony Optimization~\cite{dorigo1996ant,DORIGO2005243}. Ant-inspired swarm coordination has also been applied to foraging problems in distributed robotic systems. {Authors in \cite{hsieh2008biologically} propose a stochastic ant-inspired approach to distribute swarm agents among different target regions.} In \cite{drogoul1993some} the authors present some early experiments on how robots can lay and follow pheromones to explore a space and collect targets, and  \cite{sugawara2004foraging,fujiswarms,campo2010artificial,
alers2014insect,10.1007/978-3-030-00533-7_11} have presented similar robotic systems, either by using a \emph{digital} pheromone field \cite{campo2010artificial,
alers2014insect,10.1007/978-3-030-00533-7_11}, using real chemicals \cite{fujiswarms}, or fluorescent floors \cite{sugawara2004foraging} (see also \cite{payton2001pheromone,garnier2007alice,hrolenok2010collaborative} among others). Despite the many models proposed in entomology, and the implementations on robotic systems, little is known about the convergence guarantees of such systems. The main goal of this paper is to investigate the convergence properties of a simple version of a stigmergy-based solution to the foraging problem.

Drawing a parallelism between the pheromones of ant swarms and $Q$-values \cite{watkins1992q} one can formulate the dynamics of a stigmergy-based system as a problem that resembles traditional reinforcement learning (RL) approaches. Agents explore an environment with a set of actions to choose from, and reward (deposit pheromones) their current state (spatial location) depending on the set goal. In the works of Monekosso \cite{monekosso2001phe} a first approach was taken to mix traditional $Q$-learning and pheromone based interaction in foraging swarms. In \cite{panait2004pheromone} a variation of such utility function learning approach is presented, where the swarm uses two kinds of pheromones to distinguish between the food search and the nest search.

Nevertheless, complications arise when trying to apply $Q-$learning related strategies to study the convergence of utility-based foraging swarms. In these stigmergy-based foraging problems, solutions to the iterative utility values are coupled to the agent trajectories. This prevents us from using these sort of stochastic dynamic programming techniques to study the \emph{trajectories} of the agents in such learning stigmergy-based swarms. 
To address this problem, one can rely on the work of \cite{djarhscc,cao} about convergence of stochastic sequences of probability transition matrices. 

{Looking into stochastic systems of interacting agents, in \cite{berman2009optimized} the authors propose decentralised stochastic controllers to allocate tasks in an interacting multi-robot system given a desired target distribution. }To get rid of the stochasticity, one can look at the limiting behaviour of such systems when the number of interacting agents is taken to infinity, in what are known as \emph{mean field models}. Mean field models have been extensively used in fluid mechanics and particle physics, and more recently in game theory and control \cite{gomes2014mean,lasry2007mean}. In recent years mean field formulations of large multi-agent systems or swarms have gained increased popularity \cite{lerman2004review,elamvazhuthi2018mean,stella2018mean} (see also an extensive survey in \cite{elamvazhuthi2019mean}) as these models abstract away the stochasticity in systems where the number of interacting agents becomes very large. 

The main goals of this work are twofold: 
\begin{enumerate} 
\item First, to approximate a multi agent stochastic system for a foraging problem as a mean-field non-stochastic process. Additionally, to provide intuition on the role of the different parameters in a large multi-agent stigmergy swarm.
\item Second, to derive convergence guarantees that can be attributed to the mean field model of a stigmergy swarm, and provide insight on the resulting shape of the stationary solutions, both for the agent trajectories and the pheromone field.
\end{enumerate}

\section{Preliminaries}

\subsection{Notation}
A set whose elements depend on a parameter is indicated as $\mathcal{S}(\cdot)$. Sequences are represented as $\{A(t)\}\equiv A_t\equiv \{A(0),\,A(1),...,\,A(t)\}$, and the union of two different sequences is computed set-wise: $A_1\cup A_2:=\{i:i\in A_1\vee i\in A_2 \} $. We consider only discrete time systems, i.e. $t\in\mathbb{N}_{0}$. Unless stated otherwise, upper case letters are used for matrices ($B\in\mathbb{R}^{n\times n}$) and (bold) lower case letters for (vectors) scalars ($\mathbf{b}\in\mathbb{R}^{n}$, $b\in\mathbb{R}$). We use superscripts to distinguish between related vectors, and subscripts to indicate entries in a vector. That is, $\mathbf{a}_k^1$ is the $k-$th entry of the vector $\mathbf{a}^1$. For two vectors $\mathbf{a},\mathbf{b}\in\mathbb{R}^{n}$, we say $\mathbf{a}\geq (\leq) \mathbf{b}$ iff $\mathbf{a}_i\geq (\leq)\mathbf{b}_i \,\,\forall i$. We use $|\cdot|$ for the cardinality of a set, and $\|\cdot\|_k$ for the $k$-th norm of a vector or the $k-$th induced norm of a matrix. We define the set of all probability vectors of size $n$ as $\mathbb{P}^n:=\{\mathbf{v}\in[0,1]^n:\sum_{i=1}^n \mathbf{v}_i=1\}$. Vectors $\mathbf{1}^n,\,\mathbf{0}^n$ are the one and zero vectors of size $n$, respectively. The function $\operatorname{sgn}(\cdot)$ is the sign operator, with $\operatorname{sgn}(\mathbf{0})=\mathbf{0}$.

We say a function $f:\mathbb{R}_+\to\mathbb{R}_+$ is in the class of functions $\mathcal{K}$ if $f$ is continuous, monotonically increasing and $f(0)=0$. We say a function $f: \mathbb{R}_+\to \mathbb{R}_+$ is in class $\mathcal{K}_\infty$ if $f(\cdot)\in\mathcal{K}$ and $\lim_{a\to\infty}f(a) = \infty$.

A function assigning to each instant of time a value on each edge of a graph can be written as a matrix, and the subscript indicates both edges and entries in the image of the function. That is, let $|\mathcal{V}|$ be the number of vertices in a graph, and $f:\mathbb{N}\to\mathbb{R}^{|\mathcal{V}|\times |\mathcal{V}|}$. Then, $f_{ij}(k)$ is the $i,j$-th entry in the image $f(k)$.

When talking about stochastic processes, we use $\Omega$ as the set of outcomes in a probability space, $\mathcal{F}$ as the measurable algebra (set) of events, and $P$ as a probability function $P:\mathcal{F}\to [0,1]$. We use $E[\cdot]$ and $\operatorname{Var}[\cdot]$ for the expected value and the variance of a random variable. We say a result holds \textit{almost surely} (\textit{a.s.}) when it holds with probability $1$. When two (or more) random variables follow the same probability distribution and are independent from each other we use \textit{independent and identically distributed} (\textit{i.i.d.}).
\subsection{Weighted Graphs}
In this work we discretise geometrical (bi-dimensional) spaces using connected planar graphs. 
\begin{definition}\label{def:graph}
We define {a vertex weighted graph with time varying weights} $\mathcal{G}\coloneqq (\mathcal{V}, \mathcal{E},\mathbf{w}(t))$ as a tuple including a vertex set $\mathcal{V}$, edge set $\mathcal{E}$ and weights $\mathbf{w}:\mathbb{N}_{0} \to \mathbb{R}^{|\mathcal{V}|}_{\geq 0}$, where each value $\mathbf{w}_{i}(t)$ is the weight assigned to vertex $i\in\mathcal{V}$ at time $t$. Furthermore, the graph is connected if for every pair $i\neq j\in\mathcal{V}$ there exists a set of edges $\{(iu_1),(u_1 u_2),...\,,(u_n j)\}\subseteq\mathcal{E}$ that connects $i$ and $j$.
\end{definition}
We refer to an edge connecting $i$ to $j$ as $(ij)\equiv e\in\mathcal{E}$. Additionally, the graph is undirected if $(ij)\in\mathcal{E}\iff (ji)\in\mathcal{E}$. The adjacency matrix $A\in\mathbb{R}^{|\mathcal{V}|\times |\mathcal{V}|}$ and (out)weight matrix $W:\mathbb{N}\to\mathbb{R}^{|\mathcal{V}|\times |\mathcal{V}|}$ are
\begin{equation*}
A_{ij}\coloneqq \left\{\begin{array}{l}
1\quad \forall (ij)\in\mathcal{E},\\
0\quad \text{else.}
\end{array}\right.,\, W_{ij}(t)\coloneqq \left\{\begin{array}{l}
\mathbf{w}_j(t)\,\,\, \forall (ij)\in\mathcal{E},\\
0\quad \text{else.}
\end{array}\right.
\end{equation*}
Note that the transition weight for transition $i\to j$ is always the out-going weight $\mathbf{w}_j(t)$. For simplicity in expressions, we use the functions $D :\mathbb{R}^{n\times n}\to \mathbb{R}^{n\times n}$ and $V:\mathbb{R}^{n\times n}\to \mathbb{R}^{n\times n}$ such that $D_{ii} (B) = \sum_{j}B_{ij}, \,D_{ij} (B)=0 \,\,\forall i\neq j$, $V(B)=\operatorname{diag}\left(\max_k B_{ik}\right)$. That is, $D(B)$ is a diagonal matrix of the row sums of $B$, and $V(B)$ is a diagonal matrix containing the maximum value of every row of $B$ in the diagonal terms.
\begin{definition}\cite{diestel2012graph}
A path $p_{ij}=\mathcal{V}'\subseteq \mathcal{V}$ in $\mathcal{G}$ is any ordered subset of vertices satisfying
\begin{equation*}
\mathcal{V}'=\{i,k,l,..., z, j\}:\,\,(ik),(kl),...,(zj)\in \mathcal{E},
\end{equation*}
where no vertex appears twice. An $i$-cycle is then a path $p_{ii}$ starting and ending in the same vertex $i\in\mathcal{V}$. We refer to $\{p^k_{ij}\}$ as the set of all paths connecting $i,j$.
\end{definition}
We make use of the minimum distance between two vertices $\delta:\mathcal{V}^2\to\mathbb{N}_0^+$, $\delta(i,j):=\min_k \{|p^k_{ij}|\},$ defined only for connected pairs, and the set of minimum length paths between two vertices,
\begin{equation*}
\pi_{ij}:=\{p^*_{ij}\in\{p^k_{ij}\}:|p^*_{ij}|=\delta(i,j) \}.
\end{equation*}
The diameter of the graph is $\delta^*:=\max\{\delta(i,j)\}\quad \forall i,j\in\mathcal{V}.$ At last, it is useful to define the set of vertices in all minimum length paths as $\cup p^*_{ij} := \{v\in p^*_{ij} : p^*_{ij}\in\pi_{ij}\}$.
\subsection{Stochastic Processes and Limit Theorems}
The following definitions and theorems related to stochastic processes and random variables are used throughout this work.
\begin{definition}[Almost Sure Convergence \cite{gut2013probability}]
Let $(\Omega, \mathcal{F}, P)$ be a probability space equipped with a $\sigma$-algebra of measurable subsets of $\Omega$, with $\omega\in\Omega$ being any outcome. We say a sequence of random variables $h_0,\,h_1,...,h_t$ converges almost surely (\textit{a.s.}) to a random variable $h^*$ as $t\rightarrow\infty$ iff
\begin{equation*}
\Pr\left[\left\{\omega: h_{t}(\omega) \rightarrow h^*(\omega) \text { as } t \rightarrow \infty\right\}\right]=1.
\end{equation*}
\end{definition}

\begin{theorem}[Strong Law of Large Numbers \cite{gard1988introduction}]\label{the:lln}
Let $(\Omega, \mathcal{F}, p)$ be a probability space equipped with a $\sigma$-algebra of measurable subsets of $\Omega$. Let $h_n$ be a sequence of $n$ \textit{i.i.d.} random variables defined over the probability space, with expectation $E[h_i]$. Let $s_n=h_1 + h_2 +...+h_n$. Then,
\begin{equation*}
\lim_{n\to\infty}\frac{s_n}{n} = E[h_i]\quad \text{a.s.}
\end{equation*}
\end{theorem}
We present here a simplified version of the Perron-Frobenius theorem that we use through this work.
\begin{theorem}[Perron-Frobenius Theorem \cite{bremaud2013markov}]\label{the:pft}
Let $P\in\mathbb{R}^{n\times n}_{\geq 0}$ be a non-negative column stochastic irreducible matrix. Then,
\begin{itemize}
\item $\lambda_1(P)=1$, all other eigenvalues are smaller in norm.
\item The eigenvector $Pv=v$ defines a dimension 1 subspace with some basis vector having strictly positive entries.
\end{itemize}
\end{theorem}
At last, the following Theorem is a simplified version of results extracted from \cite{djarhscc,cao}.
\begin{theorem}[Swarm Distribution Convergence \cite{djarhscc,cao}]\label{the:hscc}
Let $P(t)$ be a column stochastic time dependent probability transition matrix with at least one odd length cycle at $t=0$, that follows stochastic dynamics. Let for some scalar $\varepsilon>0$ and $\forall \,t\geq 0$, $P_{ij}(0)>0\Rightarrow P_{ij}(t)\geq \varepsilon$. Then, the limit product
\begin{equation*}
\lim_{t\to\infty}\prod_{t_k=0}^{t}P(t_k)=\mathbf{\zeta} \mathbf{1}^T \quad \text{a.s.},
\end{equation*}
where $\mathbf{\zeta}$ is a probability vector, and does so exponentially fast with a rate no slower than $\alpha = (1-\varepsilon^{1+2\delta^*})^{\frac{1}{1+2\delta^*}}$.
\end{theorem}

\section{Foraging Swarm Model}\label{sec:1}
We state in this section the statement of a foraging problem over a graph, and present the dynamics of a proposed finite multi-agent system trying to solve the foraging problem.
\subsection{Foraging Problem}
Consider a swarm of $n$ agents moving over an undirected weighted graph $\mathcal{G}$ trying to solve a \emph{foraging problem}: the graph has a source vertex $\mathcal{S}\in\mathcal{V}$ where the agents are initialised, and a target vertex $\mathcal{T}\in\mathcal{V}$ they are supposed to find, converging to trajectories following the shortest path between $\mathcal{T}$ and $\mathcal{S}$. {\emph{Foraging} concerns, in general, both finding the shortest path between two points and depleting a food source as fast as possible. Given the discretised form of the problem, we consider in this work the \emph{foraging problem} to be solved if agents reach a state of steadily following the shortest path between $\mathcal{S}$ and $\mathcal{T}$, back and forth, since, for real agents that move at constant speed and are able to carry a limited amount of food per trip, this would be the desired scenario for maximal depletion of the food source.}

The swarm does not have accurate individual position information (GPS-like data). They can only receive measurements of a weight field from the vertices immediately next to them. Additionally, assume the agents are not able to communicate with any other member of the swarm. The agents are only able to send information to the vertex they are located at, and to receive information only from the neighbouring vertices. 
\subsection{Agent Dynamics}
We are interested in solving the foraging problem using only indirect communication through the environment (the graph). It is convenient now to introduce the assumptions that are used throughout this work.
\begin{assumption}\label{as:1}
Any undirected graph $\mathcal{G}$ is strongly connected and has at least one odd length cycle.
\end{assumption}
\begin{assumption}\label{as:1.5}
We assume there is only one $\mathcal{S}\in\mathcal{V}$ and $\mathcal{T}\in\mathcal{V}$, and the distance between them is larger than one.
\end{assumption}
\begin{remark}
Since we use graphs to discretise physical space, we can consider graphs to be triangular grids, and Assumption \ref{as:1} is always satisfied.
\end{remark}
Now, let $\mathcal{G}$ be a vertex weighted undirected graph as in Definition \ref{def:graph}. Let $A\in\{0,1\}^{|\mathcal{V}|\times |\mathcal{V}|}$ be its adjacency matrix. We define $\mathcal{A}:=\{1,2,...,n\}$ as a set of agents walking from vertex to vertex. The position of agent $a$ at time $t$ is $(t)=v,\,\,v\in\mathcal{V}$, and we group them as $x(t):=\{\mathbf{x}_a(t):\,a\in \mathcal{A}\}$. We define the vector of proportion of agents $\hat{\mathbf{q}}(t,n):\mathbb{N}_{0}\times\mathbb{N}\to\mathbb{P}^{|\mathcal{V}|}$ such that $\hat{\mathbf{q}}_i(t,n)=\frac{1}{n}|\{a\in \mathcal{A}:\mathbf{x}_a(t)=i\}|\,\,\forall i\in\mathcal{V}$. 

The position of the agents evolves depending on some probability transition matrix $P(\cdot)$. That is, for $i,j \in\mathcal{V}$,
\begin{equation}\label{ptrans}
\Pr\{\mathbf{x}_{a}(t+1)=j\,|\,\mathbf{x}_{a}(t)=i\}= P_{ji}(\cdot),\,\mathbf{x}_{a}(0)=\mathcal{S},\,\, \forall\,a\in A.
\end{equation}
In stigmergy algorithms, the transition probabilities are usually defined as the normalised weights around a vertex. In our case, drawing inspiration from experimental examples in literature, we model the probabilities of transitioning between vertices with an $\varepsilon-$greedy approach. Define the gradient matrix:
\begin{definition}\label{def:pnabla}
 Let $m_i:= |\operatorname{argmax}_k W_{ik}(t)|$. The gradient matrix $P^{\nabla}(\mathbf{w}(t))$ is a stochastic matrix such that
\begin{equation}\label{pgreed}\begin{aligned}
P^{\nabla}_{ji}(\mathbf{w}(t))&=\left\{\begin{array}{l}\frac{1}{m_i} \quad \text{if}\,\,W_{ij}=\max_k\{W_{ik}(t)\},\\
0\quad \text{else}.\end{array}\right.
\end{aligned}
\end{equation}\end{definition}
\begin{definition}
\label{def:boundedP}
Let $\mathbf{w}(t)$ be the corresponding time dependent weight matrix of a connected graph $\mathcal{G}$. Let $A$ be the adjacency matrix of the graph. For a minimum probability $\varepsilon>0$ we define the $\varepsilon-$greedy matrix as
\begin{equation*}
P^{\mathcal{G}}(t,\varepsilon):= \varepsilon (D(A)^{-1}A)^T + (1-\varepsilon)P^{\nabla}(\mathbf{w}(t)).
\end{equation*}
\end{definition}
Then, the foraging agent dynamics are as follows.
\begin{definition}
The distribution of agents $\hat{\mathbf{y}}(t)\in\mathbb{P}^{|\mathcal{V}|}$ is the probability of a given agent $a$ being on vertex $i\in\mathcal{V}$ at time $t$. This probability evolves as a random walk,
\begin{equation}\label{Py}
\hat{\mathbf{y}}_i(t+1) =\Pr\{\mathbf{x}_a(t+1)=i\}= \left(P^{\mathcal{G}}(t,\varepsilon)\hat{\mathbf{y}}(t)\right)_i\quad \forall a\in\mathcal{A},
\end{equation}
with $\hat{\mathbf{y}}_{\mathcal{S}}(0)=1\,\forall a$.
\end{definition}
The probability matrix $P^{\mathcal{G}}$ needs to be column stochastic to satisfy \eqref{Py}. Therefore, when we consider transitions $i\to j$, the corresponding probability is $P^{\mathcal{G}}_{ji}(t,\varepsilon)$ to avoid using the transposed matrix. From \eqref{Py} we define the indicator vectors 
\begin{equation}\label{zeta}\mathbf{\zeta}_i^a (t) =\left\{
\begin{array}{l}
1 \quad \text{if} \,\,\mathbf{x}_a(t) = i,\\
0 \quad \text{else,}
\end{array}\right.\,\,\hat{\mathbf{q}}(t,n) =\frac{1}{n} \sum_{a=1}^{n}\mathbf{\zeta}^a (t).
\end{equation}
Since $P^{\mathcal{G}}(t,\varepsilon)$ is the same for all agents, all $\mathbf{\zeta}^a (t)$ share the same probability distribution for all $t\geq 0$ if $\mathbf{\zeta}^a (0)=\hat{\mathbf{y}}(0) \,\,\forall a$. Then, $\hat{\mathbf{q}}(t,n)$ is a sum of identically distributed random variables with probability distribution $\hat{\mathbf{y}}(t)$.
\begin{remark}
Equation \eqref{Py} can be read as ``The probability of having an agent in some vertex $i$ at time $t+1$ is equal to the probability of being in a neighbourhood of $i$ at time $t$ times the probability of moving to $i$". However, this raises some complications. In our case $P^{\mathcal{G}}(t,\varepsilon)$ is a stochastic sequence with respect to $t$ and $P^{\mathcal{G}}(t,\varepsilon)=f(\mathcal{Q}_t(n))$. Therefore, the transition probabilities depend on the entire event history. A way of dealing with this challenge is proposed in Section \ref{sec:mf}.
\end{remark}
\subsection{Weight Dynamics}
The agents also modify the weights in the graph, similar to ants laying pheromones on the ground. Let $R(\cdot)$ be the amount of weight added to each vertex (to be properly defined below), such that $R_i(\cdot)$ is the weight added per agent to vertex $i$ at time $t$. Then, the weights in $\mathcal{G}$ evolve as
\begin{equation*}
\mathbf{w}_i(t+1)=(1-\rho)\mathbf{w}_i(t)+\rho \hat{\mathbf{q}}_i(t,n)R_i (\cdot),
\end{equation*}
where $\rho\in (0,1)$ is a chosen discount factor. The weights are initialised such that $\mathbf{w}(0)=\mathbf{1}\mathbf{w}_0$ with $\mathbf{w}_0\geq 0$. 
\begin{remark}
Keeping in mind that these systems are defined over a continuous space in reality, and to avoid over-accumulation of communication (or marking) events in one single vertex, it is useful to consider a saturated form of reinforcement, where we write \eqref{wdyn} as
\begin{equation*}
\mathbf{w}_i(t+1)=(1-\rho)\mathbf{w}_i(t)+\rho R_i(\cdot)\operatorname{sgn} \left(\hat{\mathbf{q}}_i (t,n)\right).
\end{equation*}
Effectively, this saturates the agent vector such that at every vertex there can be only one ``reinforcement" event at a given time. {From a real implementation point of view, this is logical since the reinforcement needs to be processed as some form of aggregated signal by an interacting environment or infrastructure, and otherwise such environment would need to process arbitrarily large amount of signals in finite time. Additionally, unbounded accumulation of weights may be undesirable}. From this point on, we will retain this formulation.
\end{remark}
In order for the swarm to solve the foraging problem, we draw similarities with reinforcement learning approaches to design our reward function $R$. Let $r\in\mathbb{R}_{\geq 0}$ be some positive constant, and the vector $\mathbf{\gamma}\in \mathbb{R}^{|\mathcal{V}|}_{\geq 0}$ take values
\begin{equation*}
\mathbf{\gamma}_v(r)=\left\{\begin{array}{l} r\quad \text{if}\,\,v= \mathcal{T}, \mathcal{S}\\
0\quad \text{else}.
\end{array}\right.
\end{equation*}
Then let $\lambda\in (0,1)$, and let $\Gamma (r):=\operatorname{diag}(\mathbf{\gamma}(r))$. Then, we can write the reward function in diagonal matrix form as
\begin{equation}\label{deltafvec}
R(t,r,\lambda) := \left(I+\Gamma(r) +\lambda V\left( \mathbf{w}(t)\right)\right),
\end{equation}
and the weight dynamics are simply
\begin{equation}\label{wdyn}
\mathbf{w}(t+1) = (1-\rho)\mathbf{w}(t) +\rho R(t,r,\lambda)\operatorname{sgn} \left(\hat{\mathbf{q}} (t,n)\right).
\end{equation}
The intuition about this is as follows. The reward diagonal matrix has three explicit terms in each component. First, a constant reward $1$ to all vertices to replicate the behaviour of ants: ants add pheromones to every position they are located at, with at least a minimum amount ($1$ in our case), reflecting that vertex has been visited before. Second, the term $\Gamma(r)$ where the agents reward with an additional amount $r$ the specific goals of our problem: finding $\mathcal{T}$ and returning to $\mathcal{S}$. This is also inspired in entomology; ants may mark the ground with different intensities if they have found food \cite{dussutour2009role,czaczkes2013ant}. At last, the third term is a diffusivity term (pheromones diffuse through the air to their immediate surroundings), and this term makes ants reinforce more or less based on neighbouring weights. Additionally, diffusivity is a commonly used strategy in value function learning problems. When using Q-values, diffusivity represents the maximum utility to be obtained at the next (or previous) step.

With \eqref{Py}, \eqref{zeta} and \eqref{wdyn} the stochastic dynamics of the agents and weights are fully defined. We can now present how to use such a model to obtain a foraging swarm.
\subsection{Foraging Swarm}
For the swarm to produce emerging behaviour solving the foraging problem, some additional conditions must be added into the agent behaviour and weight update rules. As proposed in several examples in the literature \cite{meyer2008tale,panait2004pheromone,payton2001pheromone}, one way to achieve this is to make use of two different pheromones (or weights $\mathbf{w}^{1}(t)$, $\mathbf{w}^{2}(t)$). {In such situation, agents looking for $\mathcal{T}$ follow $\mathbf{w}^{1}(t)$ (move according to $P^{\mathcal{G}^1}(t,\varepsilon)\equiv P^{1}(t,\varepsilon)$) and modify $\mathbf{w}^{2}(t)$, while agents looking for $\mathcal{S}$ follow $\mathbf{w}^{2}(t)$ (move according to $P^{\mathcal{G}^2}(t,\varepsilon)\equiv P^{2}(t,\varepsilon)$) and modify $\mathbf{w}^{1}(t)$}. This implies a certain ``memory" condition in the agent behaviour that may look non-Markovian: the agents follow a set of pheromones and reward another depending on their trajectory. But in fact, we can modify the system by duplicating the graph size so it remains ``memoryless".
\begin{figure}
\centering
   \resizebox{0.5\linewidth}{!}{\includegraphics{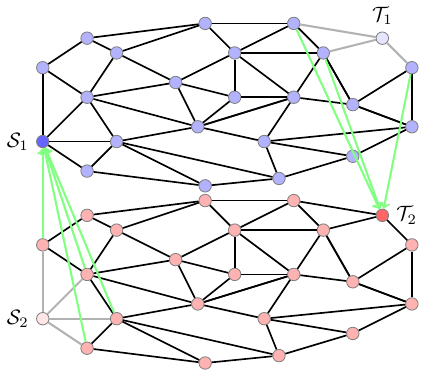}}
     \caption{Doubled interconnected Graph resulting from constructing $P(t,\varepsilon)$.}
     \label{fig:graph}
\end{figure}
The effects of this can be seen in Figure \ref{fig:graph}. In blue we have the weights $\mathbf{w}^1(t)$ and in red the weights $\mathbf{w}^2(t)$. The green edges represent the new directed edges added to the graph as a result of the interconnection of the two \emph{original} sub-graphs. The intuition behind this ``duplication" of the graph is to translate into the size of the state-space the fact that there are 2 simultaneous goals in the foraging problem (finding $\mathcal{S}$ and finding $\mathcal{T}$). We can retain the memoryless condition of the swarm by duplicating the size of the state-space and interconnecting sub-graphs. For details on how to construct this interconnected graph, see Appendix \ref{apx:graph}.

\begin{system}\label{def:forswarm}
Given two (original) undirected graphs $\mathcal{G}^{1}=\mathcal{G}^{2}=\mathcal{G}$, we define a foraging swarm {as the tuple $\phi:=(\mathcal{G},\mathcal{S},\mathcal{T},\hat{\mathbf{q}}(t,n),n,\varepsilon,\lambda)$} with $\mathcal{S}\in\mathcal{V},\mathcal{T}\in\mathcal{V},\hat{\mathbf{q}}:\mathbb{N}_{0}\times \mathbb{N}_{0}\to \mathbb{P}^{|\mathcal{V}|},n\in \mathbb{N}_{0},\varepsilon\in\mathbb{R}_{\geq 0},\lambda\in (0,1)$ such that 
\begin{equation*}\begin{aligned}
\hat{\mathbf{q}}(t,n):=& \left(\begin{array}{c}
\hat{\mathbf{q}}^{1}(t,n)\\
\hat{\mathbf{q}}^{2}(t,n)
\end{array}\right),\, \hat{\mathbf{y}}(t)\coloneqq \left(\begin{array}{l}
\hat{\mathbf{y}}^{1}(t)\\
\hat{\mathbf{y}}^{2}(t)
\end{array}\right),\\
\mathbf{w}(t)& := \left(\begin{array}{c}
\mathbf{w}^{1}(t)\\
\mathbf{w}^{2}(t)
\end{array}\right),\,W (t)\coloneqq \left(\begin{array}{cc}
W^{1}(t)& 0)\\
0&W^{2}(t)
\end{array}\right),\\
P(t,\varepsilon)&\coloneqq \left(\begin{array}{cc}
(I-T)P^{2}(t,\varepsilon)&SP^{1}(t,\varepsilon)\\
TP^{2}(t,\varepsilon)&(I-S)P^{1}(t,\varepsilon)
\end{array}\right),
\end{aligned}
\end{equation*}
initialised as $\mathbf{w}(0)=\mathbf{0}^{2\times|\mathcal{V}|}$, $\hat{\mathbf{y}}_{\mathcal{S}}(0)=1$, $\hat{\mathbf{q}}(n,0)=\mathbf{y}(0)$ which follows the dynamics
\begin{equation}\label{forswarmdyn}\begin{aligned}
\hat{\mathbf{y}}(t+1)& =P(t,\varepsilon)\hat{\mathbf{y}}(t),\\
w(t+1)& = (1-\rho)\mathbf{w}(t) +\rho R(t,r,\lambda)\operatorname{sgn} \left(\hat{\mathbf{q}}(t,n)\right),\\
R (t,r,\lambda)& :=\left(I+\Gamma(r) +\lambda V(W(t))\right)
\end{aligned}
\end{equation}
\end{system}

\begin{remark}\label{rem:disconnect}
The resulting connected graph in System \ref{def:forswarm} has some edges removed with respect to the original graph $\mathcal{G}$. It effectively disconnects $\mathcal{T}_1$ and $\mathcal{S}_2$ from the rest of the graph. Nevertheless, the density of agents initialised in these vertices is $0$, and since this is a virtual duplication of the graph, we can simply consider $\mathcal{G}\in\phi$ to have edges $\mathcal{E}= \{(ij)\in\mathcal{E}_1\cup\mathcal{E}_2:i,j\neq \mathcal{T}_1\cup\mathcal{S}_2\}$ and vertices $\mathcal{V}= \{i\in\mathcal{V}_1\cup\mathcal{V}_2:i\neq \mathcal{T}_1\cup\mathcal{S}_2\}$. This does not affect the dynamics, and results again in a strongly connected graph. We will refer to $\mathcal{T}_2\equiv \mathcal{T}$ and $\mathcal{S}_1\equiv \mathcal{S}$ as the resulting target and starting vertices in $\phi$.
\end{remark}

Observe the weight dynamics in \eqref{forswarmdyn} present coupled terms between the weights and the agents position, which is a random variable. For this reason, it becomes extremely challenging to analyse the solutions to which the system converges in such finite agent form. One way to solve this is to study what happens when we consider very large number of agents.

With the presented framework of stochastic foraging swarm, we can specify the first problem to solve in further sections.

\begin{problem*}
Let $\phi$ be a foraging swarm communicating based on a double pheromone stigmergy method. Construct a non-stochastic mean field model of the system as $n\to\infty$.
\end{problem*}

\section{Mean Field Swarm}\label{sec:mf}
In mean field models for Swarm Robotics, the number of agents is assumed to be large enough ($n\to\infty$) so that random variables can effectively be replaced by a mean valued deterministic variable. We show here how to do this in the foraging swarm presented in System \ref{def:forswarm}. Recall that the state of our system is fully defined by the $\sigma-$algebra generated by the proportion of agents in each vertex,
\begin{equation*}
\mathcal{F}_{t}=\sigma(\hat{\mathbf{q}}(0,n), \ldots \hat{\mathbf{q}}(t,n)).
\end{equation*}
Let us define the sequence $\mathcal{Q}_t(n):=\{\hat{\mathbf{q}}(0,n),...,\hat{\mathbf{q}}(t,n)\}$. In this case, an event $\mathcal{Q}_{t}(n)\in\mathcal{F}_{t}$ is a sequence of agent proportion vectors until time $t$ resulting in the generator sequence of random variables $\hat{\mathbf{q}}(0,n), \ldots \hat{\mathbf{q}}(t,n)$. Now, observe that the conditional expected value of $\mathbf{\zeta}^a(t)$ is
\begin{equation}\label{eq:expectedzeta}\begin{aligned}
E[\mathbf{\zeta}^a(t+1)=&1|\mathcal{F}_{t}]=P(t,\varepsilon)\mathbf{\zeta}^a(t).
\end{aligned}
\end{equation}
Recall that $\mathbf{\zeta}^a(0)= \hat{\mathbf{y}}(0)\,\,\forall a$, and note that while all $\mathbf{\zeta}^a (t)$ follow the same probability distribution for all $t\geq 0$ they are not independent from each other. The evolution of the probability distribution of every $\mathbf{\zeta}^a$ follows a product of probability matrices that resembles the dynamics of a Markov process. From \eqref{Py},
\begin{equation*}
\Pr[\{\mathbf{\zeta}_i^a(t)=1\}]=\hat{\mathbf{y}}_i(t)=\left(\prod_{t_k=0}^t P(t_k,\varepsilon)\hat{\mathbf{y}}(0)\right)_i.
\end{equation*}
But in this case, $P(t_k,\varepsilon)=f(\mathcal{Q}_{k})$. That is, the sequence of probability transition matrices is a function of the agent positions for all previous times. This means that, in general, for two different events $\mathcal{Q}_{t}^m(n),\mathcal{Q}_{t}^l(n)\in\mathcal{F}_t$,
\begin{equation*}
\Pr[\{\mathbf{\zeta}_i^a(t+1)=1|\mathcal{Q}_{t}^m(n)\}]\neq \Pr[\{\mathbf{\zeta}_i^a(t+1)=1|\mathcal{Q}_{t}^l(n)\}].
\end{equation*}
Furthermore, observe that the dependence is on the entire sequence until time $t$. Therefore, in general, the probability of finding agents in each vertex will depend as well on the position of other agents (making their indicator random vectors dependent). Despite this complexity, we can show convergence of the agent proportion vector to its distribution when $n\to\infty$.
\begin{theorem}\label{the:mfconv}
Let $\phi$ be a foraging swarm. Let $\mathbf{y}(t):=\lim_{n\to\infty}\hat{\mathbf{y}}(t)$, and $\mathcal{Y}_t:=\{\mathbf{y}(0),\mathbf{y}(1),...,\mathbf{y}(t)\}$ . Then,
\begin{equation*}
\mathcal{Q}^{\infty}_t:=\lim_{n\to\infty}\mathcal{Q}_t(n)=\mathcal{Y}_t\quad \text{a.s.}\quad \forall t\geq 0.
\end{equation*}
\end{theorem}
\begin{proof}[Proof (Theorem \ref{the:mfconv})]
{We show this by induction. Let us look first at $t=0$. For a fixed set of initial conditions $\hat{\mathbf{q}}(0,n)$, $\mathbf{w}(0)$, observe that $\hat{\mathbf{q}}(0,n)=\hat{\mathbf{y}}(0)$, and we have $\forall a\in\mathcal{A}$
\begin{equation*}
\Pr[\{\mathbf{\zeta}^a_i(1)=1\,|\,\hat{\mathbf{q}}(0,n),\mathbf{w}(0)\}] = \hat{\mathbf{y}}_i(1) = \left(P(0,\varepsilon)\hat{\mathbf{q}}(0,n)\right)_i.
\end{equation*}
Observe that in this case, the initial conditions are fixed, therefore we can consider the total probability 
\begin{equation}\label{eq:iid1}
\Pr[\{\mathbf{\zeta}^a_i(1)=1\} ]= \left(P(0,\varepsilon)\hat{\mathbf{q}}(0,n)\right)_i=\left(P(0,\varepsilon)\hat{\mathbf{y}}(0)\right)_i.
\end{equation}}
Observe that \eqref{eq:iid1} does not depend on $a$. Therefore, all agents have the same marginal probability distribution for $t=1$. Additionally, the transition probabilities at $t=0$ have not been affected by agent trajectories, therefore for the first time step $\mathbf{\zeta}^a_i(1)$ are \textit{i.i.d.} $\forall a$. We can then specify the joint distribution of having $k$ agents in vertex $i$ at time $t=1$: this is the joint probability of events resulting in $k$ agents moving to $i$, and $n-k$ agents moving elsewhere. Recall $\hat{\mathbf{q}}(1,n) =\frac{1}{n}\sum_{a=1}^{n}\mathbf{\zeta}^a (1).$ Since $\mathbf{\zeta}^a_i$ are indicator variables, 
\begin{equation*}
E[\mathbf{\zeta}^a(1)]= P(0,\varepsilon)\hat{\mathbf{y}}(0)=\hat{\mathbf{y}}(1).
\end{equation*}
Let us now consider the case where $n\to\infty$, and {define $\mathbf{y}(1):=\lim_{n\to\infty}\hat{\mathbf{y}}(1)$}. {Since at $t=0$ the initial conditions are fixed and all agents are initialised in the same vertex, it also holds that $\hat{\mathbf{y}}(0)=\mathbf{y}(0)$. Additionally, $P(0,\varepsilon)$ is not affected by the limit $n\to\infty$, and $\hat{\mathbf{y}}(1)=P(0,\varepsilon)\hat{\mathbf{y}}(0) =P(0,\varepsilon)\mathbf{y}(0)=\mathbf{y}(1)$.} Therefore, by Theorem \ref{the:lln} we have
\begin{equation}\label{eq:lln1}
\lim_{n\to\infty}\hat{\mathbf{q}}(1,n) = E[\mathbf{\zeta}^a(1)]=\mathbf{y}(1)\quad \text{a.s.}
\end{equation}
That is, with probability $1$, the agent proportion converges to the marginal probability distribution as $n\to\infty$ for $t=1$. From \eqref{eq:lln1} it holds that any event $\mathcal{Q}_1\in\mathcal{F}_1$ (\textit{i.e.} any possible combination of agent positions until time $t=1$) satisfies $\mathcal{Q}_1\in\mathcal{F}_1\Rightarrow \mathcal{Q}_1=\{\mathbf{y}(0),\mathbf{y}(1)\}\,\,a.s$. That is, $\Pr\{\mathcal{Q}_1\in\mathcal{F}_1:q(1)=\mathbf{y}(1)\}=1$ (the union of events has measure 1). {Then, the update of $P(1,\varepsilon)$ depends on $\mathbf{w}(1)$, and in the limit $\lim_{n\to\infty}\mathbf{w}(1)=f(\lim_{n\to\infty}\hat{\mathbf{q}}(1,n),\mathbf{w}(0))=f(\mathbf{y}(0),\mathbf{w}(0))$}. Now for $t=2$,
\begin{equation}\begin{aligned}\label{eq:lte}
E[\mathbf{\zeta}^a(2)]=&E[E[\mathbf{\zeta}^a(2)|\,\mathcal{F}_1]]=E[P(1,\varepsilon)E[\mathbf{\zeta}^a(1)|\,\mathcal{F}_1]]=\\
=&P(1,\varepsilon)E[\mathbf{\zeta}^a(1)]=P(1,\varepsilon)\mathbf{y}(1)=\mathbf{y}(2).
\end{aligned}
\end{equation}
Therefore, with probability 1, the marginal probability distributions $\mathbf{\zeta}^a(2)$ are determined by $\mathbf{y}(1)$ (since they depend on $\mathcal{Q}_1$, and this occurs \emph{a.s.}). Therefore, the variables are \textit{i.i.d.} {in the limit $n\to\infty$}, and by the law of large numbers,
\begin{equation*}
\lim_{n\to\infty}\hat{\mathbf{q}}(2,n) = E[\mathbf{\zeta}^a(2)]=\mathbf{y}(2)\quad \text{a.s.}
\end{equation*}
By induction the only possible outcome sequence is $\mathcal{Q}_t=\{\mathbf{y}(0),\mathbf{y}(1),...,\mathbf{y}(t)\}$ where $\Pr\{\lim_{n\to\infty}\mathcal{Q}_t(n)=\mathcal{Q}_t\}=1\,\,\forall t\geq 0$. Therefore $E[\mathbf{\zeta}^a(t+1)]=P(t,\varepsilon)\mathbf{y}(t)=\mathbf{y}(t+1)$, thus
\begin{equation*}
\lim_{n\to\infty}\hat{\mathbf{q}}(t,n) = \mathbf{y}(t)\quad \text{a.s.}\quad \forall t\geq 0.
\end{equation*}
\end{proof}
By making use of Theorem \ref{the:mfconv}, one can take the mean field limit and approximate the behaviour of $\hat{\mathbf{q}}(t,n)$ with $\mathbf{y}(t)$ as $n\to\infty$. Additionally, the indicator variables $\mathbf{\zeta}^a(t)$ become \textit{i.i.d.} When $n\to\infty$, there is only one possible sequence $\mathcal{Q}_t$ occurring with probability one. In other words, $\Pr[\{\mathcal{Q}_t\in\mathcal{F}_t:\mathcal{Q}_t =\mathcal{Y}_t\}]=1$, so $\mathcal{Q}_t =\mathcal{Y}_t$ happens for a set of outcomes of measure 1, and the evolution of the agent density becomes deterministic. This means that the sequence of matrices $P(t,\varepsilon)$ is also deterministic, and independent of every single $\mathbf{\zeta}^a(t)$. Therefore, the probability distribution $\mathbf{y}(t)$ of all $\mathbf{\zeta}^a(t)$ becomes independent from individual agent trajectories. This translates into the indicator vectors being \textit{i.i.d.} with respect to each other.
\begin{remark}\label{rem:yhat}
Observe the difference between $\hat{\mathbf{y}}(t)$ (probability distribution of agent positions for a finite number $n$) and  $\mathbf{y}(t)$ (probability distribution of agent positions when $n\to\infty$). In all cases, $\hat{\mathbf{y}}(0)=\mathbf{y}(0)$, but they can be different from each other for $t>0$ since they evolve according to $P(t,\varepsilon)$, which implicitly depends on $n$.
\end{remark}
We can now define our \emph{mean field swarm system}.
\begin{system}\label{def:sss2}
Let $\mathcal{G}^{1}=\mathcal{G}^{2}=\mathcal{G}$ be two identical connected weighted graphs. A mean field foraging swarm system {is defined as the tuple $\Phi:=(\mathcal{G},\mathcal{S},\mathcal{T},\mathbf{y}(t),\varepsilon,\lambda)$} with $\mathcal{S}\in\mathcal{V},\mathcal{T}\in\mathcal{V},\mathbf{y}:\mathbb{N}_{0}\to \mathbb{P}^{|\mathcal{V}|},\varepsilon\in\mathbb{R}_{\geq 0},\lambda\in (0,1)$. The state variables are initialised as $\mathbf{w}(0)=\mathbf{0}^{|\mathcal{V}|}$, $\mathbf{y}_{\mathcal{S}}(0)=1$, and:
\begin{equation}\label{sss2dyn}\begin{aligned}
\mathbf{y}(t+1)& = P(t,\varepsilon)\mathbf{y}(t),\\
\mathbf{w}(t+1)& = (1-\rho)\mathbf{w}(t) +\rho R (t,r,\lambda)\operatorname{sgn} \left(\mathbf{y}(t)\right),
\end{aligned}
\end{equation}
\end{system}

These concepts lead us to the second goal of this work.
\begin{problem*}
Let $\Phi$ be a mean field foraging swarm. Do the mean field dynamics converge to a (sub-optimal) fixed point? Additionally, what can we say (experimentally) about the deviation from the mean field case when choosing a finite number $n$?
\end{problem*}

\section{Convergence Guarantees}\label{sec:conv}
We study next the convergence properties of the mean field foraging swarm $\Phi$ of System \ref{def:sss2}.
\begin{proposition}\label{prop:yinf}
Let $\Phi$ be a mean field foraging swarm system. Let Assumption \ref{as:1} hold. {Let $1\geq\varepsilon>0$}. Then, the agent density $\mathbf{y}(t)$ converges exponentially to a stationary density $\mathbf{y}(\infty) \in\mathbb{P}^{|\mathcal{V}|}$, unique for given initial conditions $\mathbf{y}(0)$ and $\mathbf{w}(0)$, that satisfies $\mathbf{y}_i(\infty) >0\,\,\forall i\in\mathcal{V}$.
\end{proposition}
\begin{proof}
See Appendix \ref{apx:proofs}.
\end{proof}

Additionally, we can show the following result. Recall $\delta^*$ is the diameter of the underlying graph.

\begin{lemma}\label{lem:1}
Let $\Phi$ be a mean field foraging swarm system. With $t_\delta = 2\delta^*$, it holds that $\operatorname{sgn}(\mathbf{y}(t))=\mathbf{1}\quad \forall t>t_\delta.$
\end{lemma}
\begin{proof}
See Appendix \ref{apx:proofs}.
\end{proof}
Since there is a minimum probability of accessing any vertex in the graph (and the graph has odd cycles), eventually there is a non-zero amount of agents in every vertex, regardless of the foraging dynamics. Going back again to the relation finite-infinite agents, this is equivalent to saying that agents have a non-zero probability of accessing every vertex of the graph for all times greater than $t_\delta$.
\begin{remark}
In fact, $2\delta^*$ is an upper bound for the required time $t_\delta$. It represents the case where $\mathcal{G}$ is one edge away from being bipartite, and to reach some even vertex in odd time it takes $\delta^*$ time steps to reach the (only) odd length cycle plus $\delta^*$ time steps to reach the target vertex. In practice, $t_\delta\in [\delta^*,2\delta^*]$.
\end{remark}
With these preliminary results, we can present the main contribution of this section.
\begin{proposition}\label{prop:uniqueness}
There is a unique weight vector $\mathbf{w}(\infty)$ and corresponding matrix $W(\infty)$ satisfying $\mathbf{w}(\infty):=(I+\Gamma(r)+\lambda V(W(\infty)))\mathbf{1}$ for a fixed reward matrix $\Gamma(r)$ and $\lambda\in [0,1)$.
\end{proposition}
\begin{proof}
Let $B\in\{0,1\}^{|\mathcal{V}|\times|\mathcal{V}|}$ be the selector matrix satisfying $B\mathbf{w}(\infty)= V(W(\infty))\mathbf{1}$. Since $B$ is a row stochastic matrix by Theorem \ref{the:pft} it has all its eigenvalues in the unit disc, and $(I-\lambda B)$ has all its eigenvalues in a disc of radius $\lambda$ centred at 1. Therefore, its inverse is properly defined and $\mathbf{w}(\infty)=(I-\lambda B)^{-1}(I+\Gamma(r))$ has a unique solution if $\lambda\in [0,1)$.
\end{proof}
\begin{theorem}\label{the:1}
The weight dynamics in $\Phi$ have a fixed point $\mathbf{w}(\infty)$, and
\begin{equation*}
\lim_{t\to\infty} \mathbf{w}(t) = \mathbf{w}(\infty):=(I+\Gamma(r)+\lambda V(W(\infty)))\mathbf{1}.
\end{equation*}
\end{theorem}
\begin{proof}[Proof (Theorem \ref{the:1})]
Recall the weight dynamics:
\begin{equation}\label{eq:wdync1}\begin{aligned}
\mathbf{w}(t+1) =& (1-\rho)\mathbf{w}(t) + \rho\left(I+\Gamma(r)+\lambda V(W(t))\right)\operatorname{sgn}(\mathbf{y}(t)).
\end{aligned}
\end{equation}
Let $\mathbf{w}(\infty) = (I+\Gamma(r)+\lambda V(W(\infty)))\mathbf{1}$, 
$\mathbf{z}(t):= \mathbf{w}(t) - \mathbf{w}(\infty)$. Subtract $\mathbf{w}(\infty)$ at each side of \eqref{eq:wdync1}:
\begin{equation}\label{eq:12}\begin{aligned}
\mathbf{z}(t+1) =  (1-\rho)\mathbf{z}(t) + \rho\left(R (t,r,\lambda)\operatorname{sgn}(\mathbf{y}(t))-\mathbf{w}(\infty)\right).
\end{aligned}
\end{equation}
Define $\mathbf{e_y}(t):=\operatorname{sgn}(\mathbf{y}(t))-\mathbf{1}$ to obtain:
\begin{equation*}\begin{aligned}
& \left( I+\Gamma(r)+\lambda V(W(t))\right)\operatorname{sgn}(\mathbf{y}(t))-\mathbf{w}(\infty)=\\
=&(I+\Gamma(r)+\lambda V(W(t)))\mathbf{e_y}(t) +\lambda \left(V(W(t))-V(W(\infty))\right)\mathbf{1}.
\end{aligned}
\end{equation*}
Taking the $\infty$-norm at each side of \eqref{eq:12}:
\begin{equation}\label{eq:13}\begin{aligned}
\left\lVert \mathbf{z}(t+1)\right\rVert_\infty &= \left\lVert (1-\rho)\mathbf{z}(t) + \rho\left( R (t,r,\lambda)\mathbf{e_y}(t) +\lambda \left(V(W(t))-V(W(\infty))\right)\mathbf{1}\right)\right\rVert_\infty\leq\\
\leq&(1-\rho)\left\lVert \mathbf{z}(t)\right\rVert_\infty+\rho\left\lVert R (t,r,\lambda)\right\rVert_\infty \left\lVert \mathbf{e_y}(t)\right\rVert_\infty+\rho \lambda \left\lVert V(W(t))-V(W(\infty)) \right\rVert_\infty \left\lVert \mathbf{1}\right\rVert_\infty.
\end{aligned}
\end{equation}
Recall the induced $\infty$-norm of a matrix is its maximum absolute row sum. Then, 
\begin{equation}\begin{aligned}\label{eq:Vbound2}
&\left\lVert V(W(t))-V(W(\infty))\right\rVert_\infty \left\lVert \mathbf{1}\right\rVert_\infty =\max_i|\max_j \mathbf{w}_{ij}(t)-\max_j \mathbf{w}_{ij}(\infty)|\leq\\
\leq & \max_i |\max_j|\mathbf{w}_{ij}(t)-\mathbf{w}_{ij}(\infty)||=\max_i |\mathbf{z}_i(t)|=\|\mathbf{z}(t)\|_{\infty}.
\end{aligned}
\end{equation}
Now from Lemma \ref{lem:1}, $\left\lVert \mathbf{e_y}(t)\right\rVert_\infty=0\,\,\,\forall t>2\delta^*$, therefore substituting \eqref{eq:Vbound2} in \eqref{eq:13}:

\begin{equation}\label{eq:ISS}\begin{aligned}
&\|\mathbf{z}_i(t+1)\|_{\infty} \leq (1-\rho) \|\mathbf{z}(t)\|_{\infty}+\rho\lambda \|\mathbf{z}(t)\|_{\infty} =\\
&=(1-\rho(1-\lambda ))\|\mathbf{z}(t)\|_{\infty}\leq (1-\rho(1-\lambda ))^{2}\|\mathbf{z}(t-1)\|_{\infty}\leq \\
&\leq (1-\rho(1-\lambda ))^{t-2\delta^*}\|z(2\delta^*)\|_{\infty}\Rightarrow\lim_{t\to\infty} \|\mathbf{z}_i(t)\|_{\infty}=0.
\end{aligned}
\end{equation}
Finally, $\lim_{t\to\infty}\|\mathbf{z}(t)\|_{\infty}=0\Rightarrow \lim_{t\to\infty} \mathbf{w}(t) = \mathbf{w}(\infty),$ and the proof is complete.
\end{proof}
\begin{corollary}
\label{prop:eig}
The probability transition matrix converges to a unique matrix $\lim_{t\to\infty}P(t,\varepsilon)=P(\infty,\varepsilon)$, and the stationary distribution of agents $\lim_{t\to\infty}\mathbf{y}(t)= \mathbf{y}(\infty) $ is the eigenvector corresponding to the eigenvalue 1. That is,
\begin{equation}
P(\infty,\varepsilon):=\lim_{t\to\infty}P (t,\varepsilon),\,\,\,\, \mathbf{y}(\infty)  = P(\infty,\varepsilon) \mathbf{y}(\infty) .
\end{equation}
\end{corollary}
\begin{proof}
See Appendix \ref{apx:proofs}.
\end{proof}

\subsection{On the optimality of solutions}
Let us examine now what do the agent distributions look like in a mean field swarm system $\Phi$. To this end, we define first a few useful concepts to characterize the state variables. 

{\begin{definition}
Let $\mathbf{w}$ be the weight vector in a system $\Phi$. We define a maximum (weight) gradient set of paths $\pi_{ij}^{\nabla }(\mathbf{w})$ as the set of all unique paths between vertices $i,\,j$ satisfying
\begin{equation*}\begin{aligned}
&{p_{ij}^{\nabla }\in \pi_{ij}^{\nabla }(\mathbf{w})} \iff p_{ij}^{\nabla }:=\{i,i_2,i_3,...,i_k,j\},\\
&i_2=\operatorname{argmax}_v(W_{iv}(t)),\,\,i_3=\operatorname{argmax}_v(W_{i_2 v}(t)),...,\\
&j=\operatorname{argmax}_v(W_{i_k v}(t)).
\end{aligned}
\end{equation*}
\end{definition}}
In other words, let the weight vector be $\mathbf{w}(t)$. Then, $\pi_{ij}^{\nabla }(\mathbf{w}(t))$ is the set of all paths obtained from following the maximum neighbouring weights at each step when going from $i$ to $j$.
Note that, for any two $i,j\in\mathcal{V}$, it can be that $\pi_{ij}^{\nabla }(\mathbf{w}(t))=\emptyset$ if picking the maximum weight neighbour at every step does never connect $i$ with $j$.
\begin{definition}\label{def:optW}
We define the set of optimal weight vectors $\mathcal{W}^*\subset \mathbb{R}^{|\mathcal{V}|}_{\geq 0}$ for a mean field foraging system $\Phi$ as 
\begin{equation*}\begin{aligned}
\mathcal{W}^*:=\{\mathbf{w}^*:&\pi_{ij}^{\nabla }(\mathbf{w}^*)\equiv\pi_{ST}\wedge \pi_{ij}^{\nabla }(\mathbf{w}^*)\equiv\pi_{TS}\}.
\end{aligned}
\end{equation*}
That is, for every weight vector $\mathbf{w}^*\in \mathcal{W}^*$, the set of paths resulting from starting at $\mathcal{S}\,(\mathcal{T})$ and following the maximum gradient vertices lead to $\mathcal{T}\,(\mathcal{S})$, and is equal to $\pi_{ST}\,(\pi_{TS})$.
\end{definition}
This interpretation of an optimal set of weights is entirely pragmatic. We call a weight distribution optimal if, when starting at $\mathcal{S}$ and following the maximum weight vertex at every step, we end up at $\mathcal{T}$ and we obtain a minimum length path between the two (and vice-versa from $\mathcal{T}$ to $\mathcal{S}$). {Additionally, observe that the optimal weight set $\mathcal{W}^*$ is defined for the \emph{doubled} graph in Figure \ref{fig:graph}. Nevertheless, given the symmetry of the graph (the sub-graphs satisfy $\mathcal{G}^1=\mathcal{G}^2$), any optimal weight vector $\mathbf{w}^*\in \mathcal{W}^*$ generates optimal paths on the original (unweighed) graph too, but it does so separately for paths $\mathcal{S}\to\mathcal{T}$ and for paths $\mathcal{T}\to\mathcal{S}$.} Intuitively, constructing a weight vector $\mathbf{w}^*$ means the swarm has solved the foraging problem by building a weight function whose gradient always leads towards an optimal path. 
\begin{proposition}\label{cor:1}
Let $\Phi$ be a mean field stigmergy swarm. Then, $\lim_{t\to\infty}\mathbf{w}(t)=\mathbf{w}(\infty)\in \mathcal{W}^*$.
\end{proposition}
\begin{proof}
See Appendix \ref{apx:proofs}.
\end{proof}
From Definition \ref{def:boundedP}, abusing the notation for the variable $\varepsilon$ we can decompose $P (\infty,\varepsilon)$ in two matrices such that
\begin{equation}\label{eq:pepsilon}
P (\infty,\varepsilon)=(1-\varepsilon)P(\infty,0) +\varepsilon P(\infty,1),
\end{equation}
where $P(\infty,0)$ is the transition matrix corresponding to moving according to the gradient of the weights $\mathbf{w}(\infty)$. Observe as well that $P(\infty,1)$ depends only on the adjacency matrix $A$, which guarantees the decomposition \eqref{eq:pepsilon} to be unique. We define then the following sets.
\begin{definition}
We define $N_v^{out}= \{j\in\mathcal{V}:P_{jv} (\infty,0)>0\}$, and $N_v^{in}= \{j\in\mathcal{V}:P_{vj} (\infty,0)>0\}$ as the out and in neighbour vertices connected to $v$ by following $P(\infty,0)$.
\end{definition}
Observe that in Definition \ref{def:pnabla} we use $m$ to count the number of (out) neighbours that have maximum weight around a vertex, and therefore $m_v\equiv |N_v^{out}|$. Now let $k=\delta(\mathcal{S},\mathcal{T})$, and recall $p^*_{ST}\in\pi_{ST}$ is any path in the set of optimal paths. Let $\overline{\mathbf{y}}\in\mathbb{P}^{|\mathcal{V}|}$ be a probability vector taking values
\begin{equation*}\begin{aligned}
\overline{\mathbf{y}}_i :=&\left\{\begin{array}{l}\frac{1}{2k}\quad\quad\quad\quad\quad\quad\quad\quad\quad\quad\,\,\, \text{if} \,\,i=\mathcal{S},\mathcal{T},\\
\frac{1}{2k}\sum_{p\in\pi_{Si}}\prod_{u\in p\setminus i}\frac{1}{|N_u^{out}|} \quad \text{if} \,\,i\in \cup p^*_{ST}\setminus \mathcal{S},\mathcal{T},\\
\frac{1}{2k}\sum_{p\in\pi_{Ti}}\prod_{u\in p\setminus i}\frac{1}{|N_u^{out}|} \quad \text{if} \,\,i\in\cup p^*_{TS}\setminus \mathcal{S},\mathcal{T},\\
0\quad\quad\quad\quad\quad\quad\quad\quad\quad\quad\quad \text{else}.
 \end{array} \right.
\end{aligned}
\end{equation*} The term $\frac{1}{|N_u^{out}|}$ can be interpreted as the probability of moving out of $u$ towards a specific neighbour, therefore the product $\Pr\{p\}:=\prod_{u\in p\setminus i}\frac{1}{|N_u^{out}|}$ can be interpreted as the probability of following a path $p$ until vertex $i$, starting at $\mathcal{S},\mathcal{T}$. Then, we obtain the following result.
\begin{proposition}\label{prop:optimal} {Let $\Phi$ be a mean field stigmergy swarm. Let the system converge as $t\to\infty$ for a fixed $1>\varepsilon>0$ and let $P(\infty,0)$ defined in \eqref{eq:pepsilon}. Then, $P(\infty,0)\overline{\mathbf{y}}=\overline{\mathbf{y}}.$}
\end{proposition}
\begin{proof}
See Appendix \ref{apx:proofs}.
\end{proof}
{Proposition \ref{prop:optimal} indicates that the vector $\overline{\mathbf{y}}$ is the first eigenvector of the ``gradient" matrix $P(\infty,0)$. That is, as the weights converge, when the agents move by selecting the maximum weight vertex around them, the only stationary distribution is the one that spreads all agents equally across the optimal paths between $\mathcal{S}$ and $\mathcal{T}$.}

\begin{theorem}\label{lem:kinf} Let $\Phi$ be a mean field stigmergy swarm. Let $\beta:[0,\infty)\to [0,\infty)$ be $\beta\in\mathcal{K}_\infty$. Then, it holds that
\begin{equation*}
\lVert \mathbf{y}(\infty)-\overline{\mathbf{y}} \rVert_1\leq \beta(\varepsilon).
\end{equation*}
That is, the stationary agent distribution of $\Phi$ gets arbitrarily close to the optimal distribution as $\varepsilon\to 0$.
\end{theorem}
\begin{proof}[Proof (Theorem \ref{lem:kinf})]
Recall $P (\infty,\varepsilon)=(1-\varepsilon)P(\infty,0) +\varepsilon P(\infty,1).$
Additionally, from Corollary \ref{prop:eig} and Proposition \ref{prop:optimal},
\begin{equation*}
P(\infty,\varepsilon)\mathbf{y}(\infty)=\mathbf{y}(\infty),\quad P(\infty,0)\overline{\mathbf{y}}=\overline{\mathbf{y}}.
\end{equation*}
Now let $L:=\left(I-P(\infty,0)\right)$, $\Delta P :=P(\infty,1)-P(\infty,0)$. Then, we can expand
\begin{equation}\label{eq:Lbound}\begin{aligned}
&\mathbf{y}(\infty)-\overline{\mathbf{y}} = P(\infty,\varepsilon)\mathbf{y}(\infty)-P(\infty,0)\overline{\mathbf{y}}=\\
=&(1-\varepsilon)P(\infty,0)\mathbf{y}(\infty) +\varepsilon P(\infty,1)\mathbf{y}(\infty)-P(\infty,0)\overline{\mathbf{y}}=\\
=&P(\infty,0)(\mathbf{y}(\infty)-\overline{\mathbf{y}})+\varepsilon\Delta P \mathbf{y}(\infty)\Rightarrow L(\mathbf{y}(\infty)-\overline{\mathbf{y}})=\varepsilon\Delta P \mathbf{y}(\infty).
\end{aligned}
\end{equation}
The null space of $L$ is given by $Lv=0\iff P(\infty,0)v=v$, and by Theorem \ref{the:pft} we know $v$ is unique, therefore $\operatorname{rank}(L)=|\mathcal{V}|-1$. But to solve the system of equations $L(\mathbf{y}(\infty)-\overline{\mathbf{y}})=\varepsilon\Delta P \mathbf{y}(\infty)$, $L$ needs to be invertible. For this we can add the following additional equation: We know it must hold that $\mathbf{1}^T (\mathbf{y}(\infty)-\overline{\mathbf{y}})=0$, and this equation is linearly independent from all rows in $L$ if and only if $\nexists\mu \in\mathbb{R}^{|\mathcal{V}|}$ that satisfies $L\mu = \mathbf{1}$. Let us show that there does not exist such a $\mu$ by contradiction. Assume $\exists\mu: L\mu = \mathbf{1}$. Recall $\pi_{ST},\pi_{TS}$ are the sets of optimal paths between $\mathcal{S},\mathcal{T}$ and $\mathcal{T},\mathcal{S}$, with $p^*_{ST}\in\pi_{ST}$. Then, $\forall i_1\in N_\mathcal{S}^{out}$,$L_{\mathcal{S}\mathcal{S}}=L_{i_1 i_1}=1,\quad L_{i_1\mathcal{S}}=-\frac{1}{|N_\mathcal{S}^{out}|}.$
Adding the rows of $L$ $\forall i_1$:
\begin{equation}\label{eq:sum1}
\left(\sum_{i_1\in N_\mathcal{S}^{out}}L_{i_1}\right)_j = \left\{\begin{array}{l}1\quad \text{if}\,j\in N_\mathcal{S}^{out},\\
-1\quad \text{if}\,j=\mathcal{S},\\
0\quad \text{else}.
\end{array}\right.
\end{equation}
Now let $\cup_{i_1} N_{i_1}^{out}:=\{k:k\in N_{i_1}^{out}\,\forall\, i_1\in N_\mathcal{S}^{out}\}$ be the set of all vertices at distance 2 from $\mathcal{S}$ when following optimal paths, and $i_2\in\cup_{i_1} N_{i_1}^{out}$. Adding the rows of $L$ $\forall i_2$:
\begin{equation}\label{eq:sum2}
\left(\sum_{i_2\in\cup_{i_1}  N_{i_1}^{out}}L_{i_2}\right)_j = \left\{\begin{array}{l}1\quad \text{if}\,j\in \cup_{i_1}  N_{i_1}^{out},\\
-1\quad \text{if}\,j\in N_\mathcal{S}^{out},\\
0\quad \text{else}.
\end{array}\right.
\end{equation}
Now it is clear that adding \eqref{eq:sum1} and \eqref{eq:sum2}:
\begin{equation}
\left(\sum_{i\in N_\mathcal{S}^{out}}L_{i_1}+\sum_{i_2\in\cup_{i_1} N_{i_1}^{out}}L_{i_2}\right)_j = \left\{\begin{array}{l}1\quad \text{if}\,j\in \cup_{i_1} N_{i_1}^{out},\\
-1\quad \text{if}\,j=\mathcal{S},\\
0\quad \text{else}.
\end{array}\right.
\end{equation}
Extending the sum until vertex $\mathcal{T}$, we add rows $\forall i\in\cup p^*_{ST}$:
\begin{equation}\label{eq:sum4}
\left(\sum_{i\in \cup p^*_{ST}\setminus \mathcal{S}} L_{i}\right)_j=\left\{\begin{array}{l}1\quad \text{if}\,j=\mathcal{T},\\
-1\quad \text{if}\,k=\mathcal{S},\\
0\quad \text{else}.
\end{array}\right.
\end{equation}
Analogously, considering the reverse paths $\pi_{TS}$ one obtains
\begin{equation}\label{eq:sum42}
\left(\sum_{i\in \cup p^*_{TS}\setminus \mathcal{T}} L_{i}\right)_j=\left\{\begin{array}{l}1\quad \text{if}\,j=\mathcal{S},\\
-1\quad \text{if}\,j=\mathcal{T},\\
0\quad \text{else}.
\end{array}\right.
\end{equation}
Define $\theta: = |\cup p^*_{ST}|=|\cup p^*_{TS}|$ as the number of vertices in all optimal paths, and from \eqref{eq:sum4} and \eqref{eq:sum42} one obtains
\begin{equation*}\begin{aligned}
&\sum_{i\in \cup p^*_{ST}\setminus \mathcal{S}} L_{i}\mu =\theta-1,\sum_{j\in \cup p^*_{TS}\setminus \mathcal{T}} L_{j}\mu=\theta-1\Rightarrow  -\theta= \theta,
\end{aligned}
\end{equation*}
which is a contradiction. Then, $\nexists \mu:L\mu=\mathbf{1}$, and there is a row in $L$, $\Delta P$ such that replacing it (assuming it is the last row, without loss of generality) we obtain

\begin{equation*}
\tilde{L} := \left(\begin{array}{c}L_1\\
...\\
\mathbf{1}^T
\end{array}\right), \tilde{\Delta P }\mathbf{y}(\infty) := \left(\begin{array}{c}\Delta P_1 \mathbf{y}(\infty)\\
...\\
\mathbf{0}^T
\end{array}\right),
\end{equation*}
where $\operatorname{rank}(\tilde{L})=|\mathcal{V}|$. Now, observe
\begin{equation*}
\tilde{L}(\mathbf{y}(\infty)-\overline{\mathbf{y}})=\varepsilon\tilde{\Delta P} \mathbf{y}(\infty)\Rightarrow \mathbf{y}(\infty)-\overline{\mathbf{y}}=\varepsilon\tilde{L}^{-1}\tilde{\Delta P} \mathbf{y}(\infty).
\end{equation*}
At last, since $\|\tilde{L}^{-1}\|_1$ is bounded and does not depend on $\varepsilon$ and $\|\tilde{\Delta P} \mathbf{y}(\infty)\|_1\leq 2$, $\exists c\in\mathbb{R}_{\geq 0}$ and $ \beta(\varepsilon)\in\mathcal{K}_{\infty}$ such that
\begin{equation}\begin{aligned}
\|\mathbf{y}(\infty)-\overline{\mathbf{y}}\|_1\leq\varepsilon\|\tilde{L}^{-1}\tilde{\Delta P} \mathbf{y}(\infty)\|_1\leq \varepsilon c =:\beta(\varepsilon).
\end{aligned}
\end{equation}
\end{proof}
{ We can now reflect on the similarities between a Q-Learning (or other value function iteration) strategy for finding optimal policies on a MDP and our weight-based foraging problem, as discussed in the introduction. There is a parallelism between the Q values associated to state-action pairs and the weight values (pheromones) associated to vertices on our graph: In both cases they represent the ``utility" of a state. However, in our case, by taking the mean field limit we can study the limit distribution of agents interacting with this utility field, as well as the utility values themselves. Additionally, in the mean field limit we can derive deterministic guarantees about both the \emph{distribution} of agents around the graph (i.e. the distance $\lVert \mathbf{y}(\infty)-\overline{\mathbf{y}} \rVert_1$) and the \emph{trajectories} of the agents given by the matrix $P(\infty,\varepsilon)$.}
\section{Experiments}\label{sec:exp}
Some experiments are here presented to verify the results presented in Section \ref{sec:conv}. All experiments were performed over a $20\times 20$ triangular lattice graph, which has $\min_i g_i^{out}=2$, $\max_i g_i^{out}=6$ and $\delta^* = 31$. The parameters used are presented in Table \ref{tab:params}. It is worth mentioning that the amount of ``parameter tuning" applied is  minimal. The guarantees from Section \ref{sec:conv} ensure the mean field process converges to (a neighbourhood of) a set of vertices along the shortest path, and we choose parameters to obtain representative results when having a finite amount of agents in the graph. We picked $\lambda\approx 1$ to have high diffusion, $r=5$ to be significantly higher than the unitary reinforcement in $R(t,r,\lambda)$, $\varepsilon=0.5$ to have an average exploration rate and $\rho=0.005$ since this yields an evaporation of $(1-\rho)^{4\delta^*} \approx 1/2$.
\begin{table}[h]\centering
\begin{tabular}{ |c|c|c|c| }
  \hline
  \multicolumn{4}{|c|}{Parameters} \\
  \hline
  $\rho$ & $\lambda$&$r$& $\varepsilon$\\
  \hline
  $0.005$ & $0.9$ & $5$ & $0.5$ \\
  \hline
\end{tabular}
\vspace{2mm}
\caption{\label{tab:params} Simulation parameters}
\end{table}
\subsection{Mean Field Process}
\begin{figure}[!t]
\centering
\begin{subfigure}[t]{.48\textwidth}
\centering
   \includegraphics[width=\textwidth]{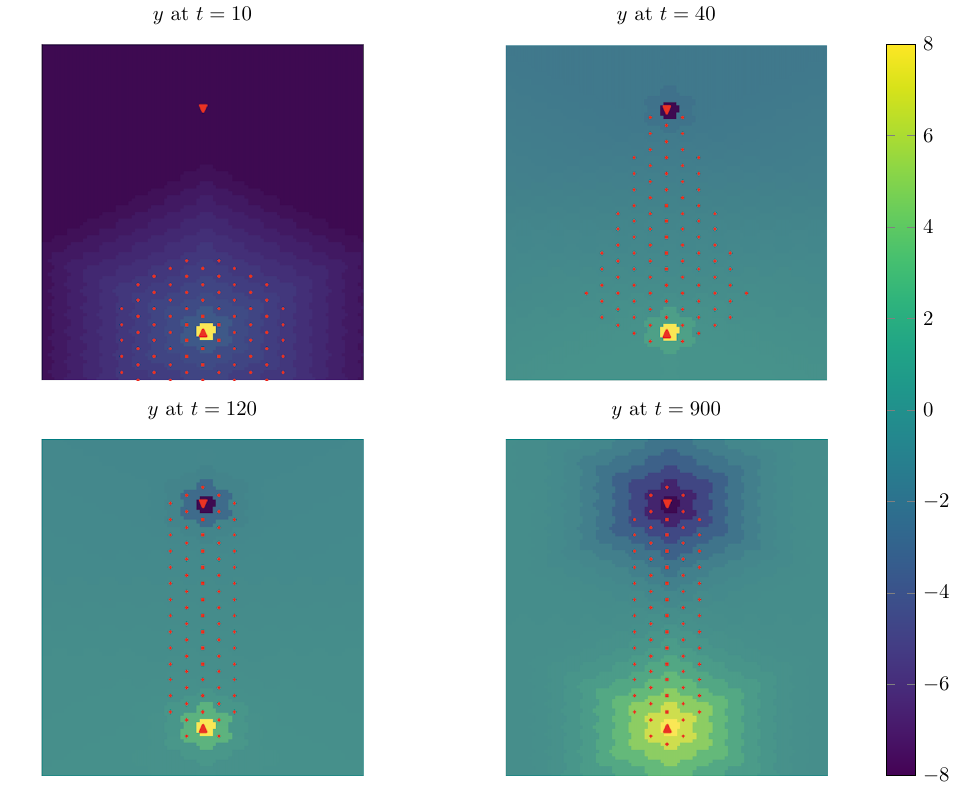}
     \caption{Mean Field results without obstables}
     \label{fig:y}
\end{subfigure}
\begin{subfigure}[t]{.48\textwidth}
\centering
   \includegraphics[width=\textwidth]{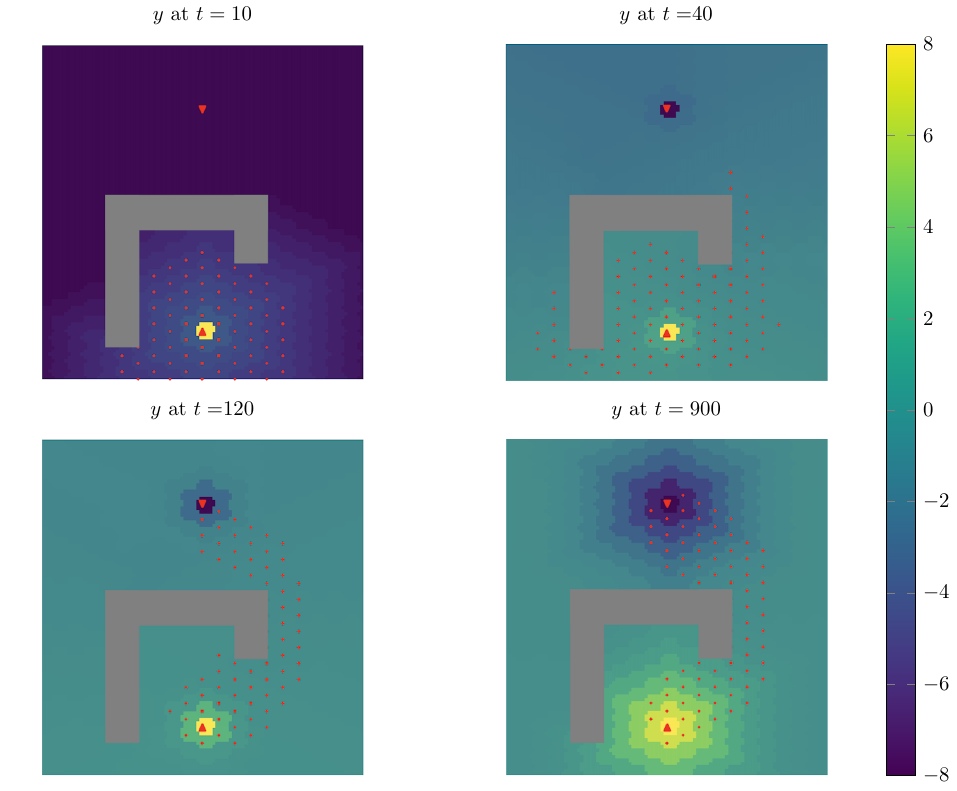}
     \caption{Mean Field results with obstacles}
     \label{fig:yobs}
\end{subfigure}
\label{fig:X}
\caption{Mean Field Results}
\end{figure}

In Figures \ref{fig:y} and \ref{fig:yobs} we present the results for two different scenarios of simulations for a mean field system $\Phi$:
\begin{enumerate}
\item One without obstacles where the shortest path is a perfect line between nest (red triangle) and food source (upside red triangle).
\item One with a sample (non-convex) obstacle where the shortest path (or collection of paths) has to go around it on the right side.
\end{enumerate}
At every vertex we plot the value $\mathbf{w}^1_i(t)-\mathbf{w}^2_i(t)$, with the color bar representing the values of the last plot. In this way we can see the vertices that have a higher overall weight corresponding to each goal. The number of agents is then proportional to the size of the red markers on the vertices. The behaviour of the system is specially interesting in the case of the obstacles in Figure \ref{fig:yobs}. In the first few time-steps there is a random exploration taking place, and very early (around $t=40$) the agent density starts accumulating in the diagonal line outwards from the nest, indicating that the shortest path is starting to be exploited. Soon after (around $t=120$) the shortest connecting path can already be observed, but the agent distribution presents oscillations. After enough time-steps the oscillations dampen (since the graph has enough odd length cycles) and the distribution converges to $\mathbf{y}(\infty)$. It is important to remark the convergence speed of the mean field dynamics; $\mathcal{G}$ has $452$ vertices, and the agent density has converged to the shortest path in about $300$ time-steps.

\begin{figure}
   \resizebox{0.5\linewidth}{!}{\includegraphics{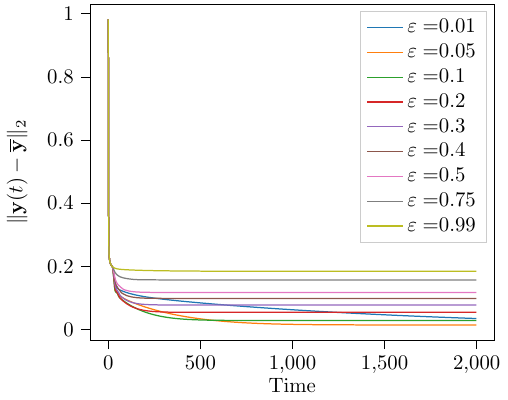}}
     \caption{Mean Field trajectories compared to $\overline{\mathbf{y}}$}
     \label{fig:ybar}
\end{figure}

In Figure \ref{fig:ybar} we present the temporal trajectories for the mean field system compared to the optimal vector $\overline{\mathbf{y}}$, for different values of $\varepsilon$. Plots for different $\rho$ and $\lambda$ values are not included since these parameters did not have an impact on the mean field results. To better isolate the effect of every parameter in the system, the influence of $\varepsilon$ is only studied on the mean field system, and later a fixed value of $\varepsilon$ is applied to the rest of the experiments.

\subsection{Finite Agents vs. Mean Field}
\begin{figure}[!t]
\centering
\begin{subfigure}[t]{.48\textwidth}
\centering
   \includegraphics[width=\textwidth]{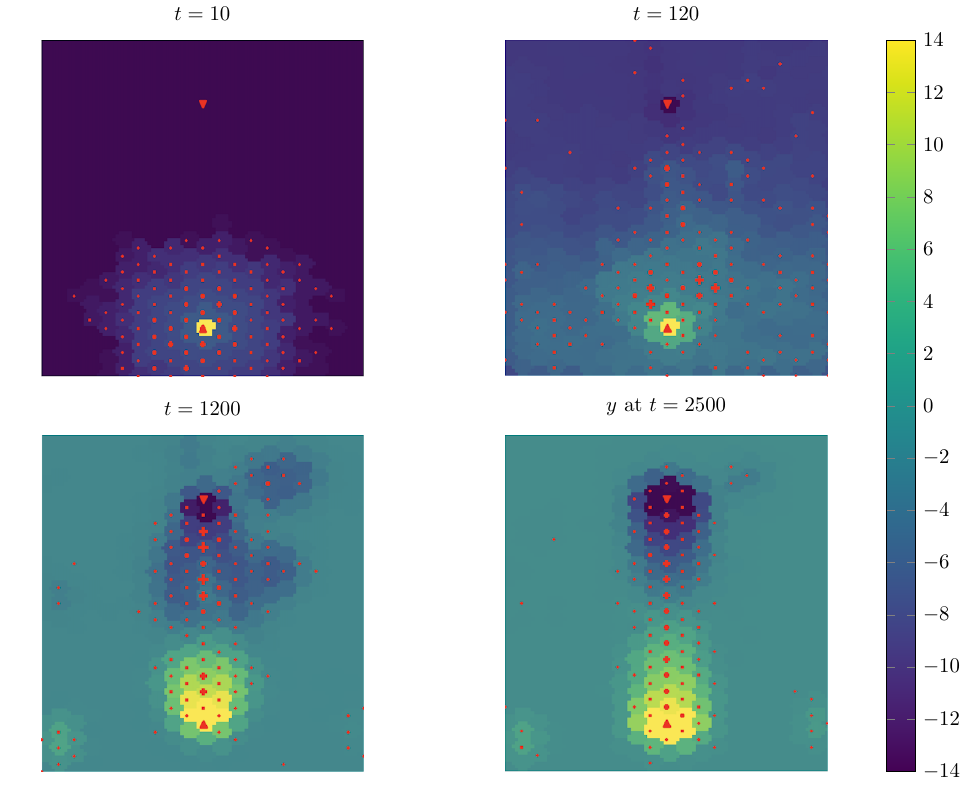}
     \caption{Discrete agent swarm with $n=600$}
     \label{fig:a}
\end{subfigure}
\begin{subfigure}[t]{.48\textwidth}
\centering
   \includegraphics[width=\textwidth]{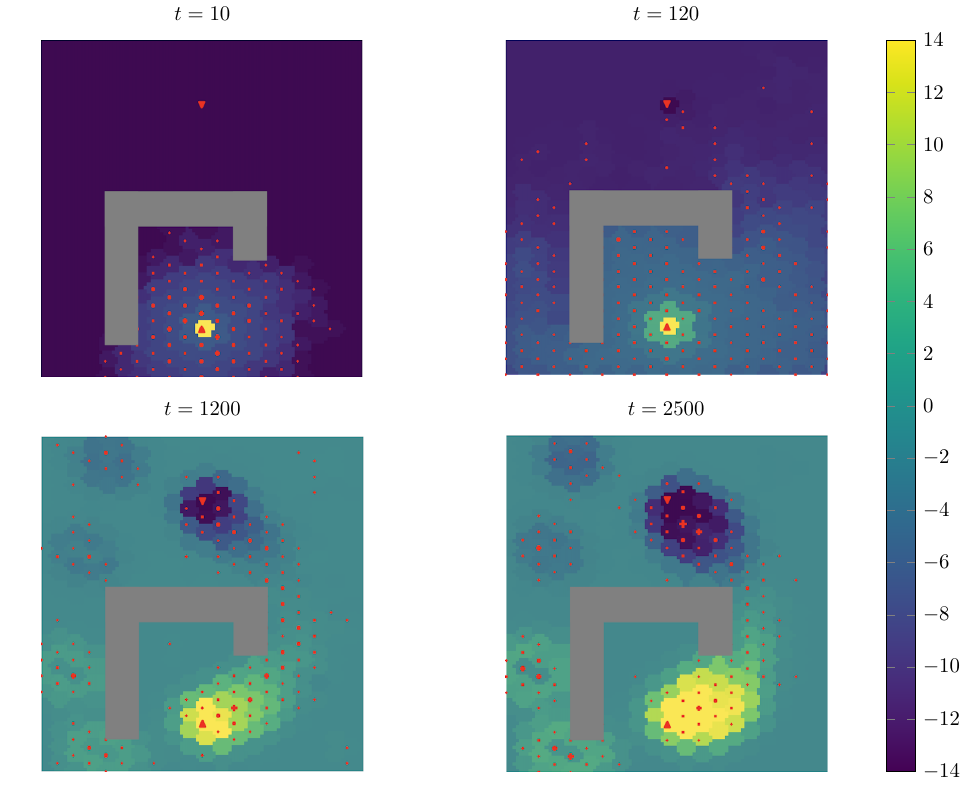}
     \caption{Discrete agent swarm with $n=600$ and obstacles}
     \label{fig:aobs}
\end{subfigure}
\label{fig:Z}
\caption{Finite agent results}
\end{figure}

We compare now the results obtained from a mean field approximation system to the ones obtained when using a finite number of agents. Figures \ref{fig:a} and \ref{fig:aobs} show a similar scenario from the mean field case, but in this case for a finite number of agents. As it can be seen, both cases take longer to achieve convergence to the shortest path. Additionally the agents concentrate over wider regions, and some are ``trapped" in irrelevant parts of the graph. Other examples in literature \cite{panait2004pheromone} solve this by re-setting the agent position if they have not found the goal vertices over a too long period of time.

\begin{figure}[!t]
\centering
\begin{subfigure}[t]{.48\textwidth}
\centering
   \includegraphics[width=\textwidth]{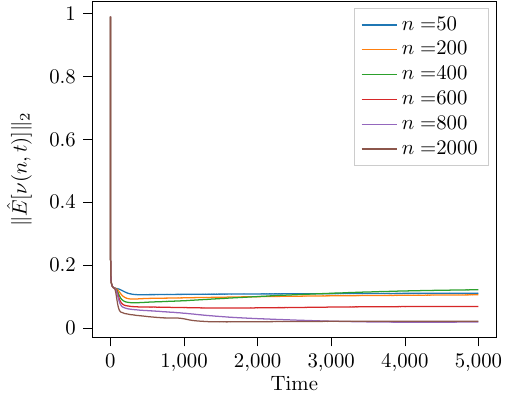}
     \caption{Sample expectation of $\nu(t,n)$ for different $n$}
     \label{fig:ek}
\end{subfigure}
\begin{subfigure}[t]{.48\textwidth}
\centering
   \includegraphics[width=\textwidth]{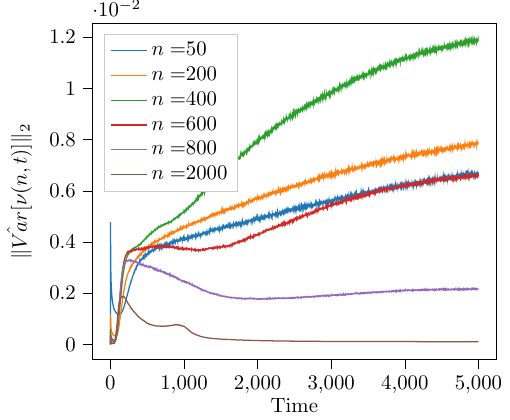}
     \caption{Sample variance of $\nu(t,n)$ for different $n$}
     \label{fig:vark}
\end{subfigure}
\label{fig:W}
\caption{Expected value and variance results}
\end{figure}

We now study the impact of having a reduced number of agents compared to the optimal solutions obtained in the mean field case. Let us for this define an error random variable:
\begin{equation*}
\nu(t,n):=\hat{\mathbf{q}}(t,n)-\mathbf{y}(\infty),
\end{equation*}
and, finite sample expectation and variance as 
\begin{equation*}\begin{aligned}
&\hat{E}[\nu(t,n)]:=\frac{1}{K}\sum_{k=1}^K \nu(t,n),\\ 
&\hat{\operatorname{Var}}[\nu(t,n)]:=\frac{1}{K}\sum_{k=1}^K(\nu(t,n)-\hat{E}[\nu(t,n)])^2.
\end{aligned}
\end{equation*}
In Figures \ref{fig:ek} and \ref{fig:vark} show the results over $K=5000$ runs. As expected from Theorem \ref{the:mfconv}, both the mean and the variance approach zero for large times as $n$ increases. Interestingly, they both exhibit a peak value after a few time-steps into the runs. This is likely due to the fact that when agents find $\mathcal{T}$ the weights change quite fast since reward is added to $\mathcal{G}^2$ suddenly, and the stochastic system runs may be prone to differ more from each other.
\subsection{Interpretation of Variance results}
Figures \ref{fig:ek} and \ref{fig:vark} show the norms of the finite sample (error) expectation and the finite sample variance. As expected, for large numbers of agents the plots go to zero relatively quickly. However, it is interesting to note that the variance and expectation of error increase with $n$ until $n\approx 600$. A possible justification for this is that there is a threshold under which more agents cause more disorder, but not necessarily better solutions. Looking at the variance values at $t=5000s$, the first curve for $n=50$ settles around $0.6\cdot 10^{-2}$, and the following curves for $n=200$, $n=400$ go up until around $10^{-2}$. This indicates that the variance increases for a range of $n$ values, until a certain threshold where it decreases until approaching $0$ for $n>1000$. 
\begin{table}[h]\centering
\begin{tabular}{ |c|c|c|c|c|c|c| }
\hline
 $r$& $\rho$ & $n$ & $\|\hat{E}[\nu(n,\overline{t})]\|_2$ & $\|\hat{\operatorname{Var}}[\nu(n,\overline{t})]\|_2$\\
\hline
$20$ & $0.005$ & $200$ &\cellcolor{blue!46} $0.112$ & \cellcolor{olive!34}$0.0097$ \\
  \hline
$5$ & $0.05$ & $200$ & \cellcolor{blue!50}$0.121$ & \cellcolor{olive!50}$0.0140$ \\
  \hline
 $5$ & $0.005$ & $200$ &\cellcolor{blue!43} $0.105$ & \cellcolor{olive!28}$0.0078$ \\
  \hline
     $20$ & $0.005$ & $800$ & \cellcolor{blue!15}$0.028$ & \cellcolor{olive!4}$0.0006$ \\
  \hline
 $5$ & $0.05$ & $800$ & \cellcolor{blue!10}$0.022$ & \cellcolor{olive!2}$0.0003$ \\
  \hline
 $5$ & $0.005$ & $800$ & \cellcolor{blue!7}$0.018$ & \cellcolor{olive!8}$0.0021$ \\
  \hline
\end{tabular}
\vspace{2mm}
\caption{\label{tab:results} Simulation results}
\end{table}

Table \ref{tab:results} displays the end results of different combination of parameters over $5000$ runs, for $\overline{t} = 5000$ and fixed $\varepsilon = 0.5$, $\lambda = 0.99$. {In general, lower $\rho$ values and larger agent numbers seem to cause smaller variances and smaller $\nu(t,n)$ values. However, for large swarms ($n=800$) decreasing the evaporation results in an increase in variance. This effect seems to be caused by the fact that for large enough swarms, higher evaporation actually pulls agents towards the optimal solutions faster, therefore decreasing the variance (or diversity) in trajectories.} Interestingly, the impact of $r$ in $\nu(t,n)$ seems to be small for the tested cases. Further study of this issue is left for future work, since it may have implications on other multi-agent stochastic systems where stochastic processes exhibit couplings that vanish for large number of agents.
\section{Discussion}
We have shown throughout this work how a multi-agent collaborative system solving a foraging problem can be approximated by a mean field formulation of the problem when $n\to \infty$. In section \ref{sec:conv} we developed formal results on the convergence and optimality of such mean field foraging system. We are able now to draw a set of conclusions from these results, combined with the experiments in section \ref{sec:exp}. 

First, the mean field foraging system converges to a unique stationary solution, and does so exponentially fast, under the proposed conditions. In fact, the distance between the mean field agent distribution $\mathbf{y}(\infty)$ and the optimal distribution $\overline{\mathbf{y}}$ seems to only depend on the exploration rate $\varepsilon$ (see Figure \ref{fig:ybar}, Theorem \ref{lem:kinf}). {That is, the evaporation (or learning) rate $\rho$ and the discount factor $\lambda$ do not have an effect in the stationary solutions, nor in the convergence speed of the mean field system. This can be explained by the fact that $\rho$ and $\lambda$ act as scaling parameters that do not change the shape of the weight gradients, thus not having an impact on the matrices $P(t,\varepsilon)$.} Note as well from the results in Figure \ref{fig:ybar} that the distance between $\mathbf{y}(t)$ and $\overline{\mathbf{y}}$ shows some linearity with $\varepsilon$ as obtained in the bounds of Theorem \ref{lem:kinf}. This indicates that, for collaborative multi-agent systems in stochastic settings, learning rates and discount factors cease to have an impact when considering large numbers of agents. Therefore, the study of mean field limits on such a multi-agent system allows us to de-couple the influence of some parameters, that may only come into play when considering small agent numbers.

Second, lower exploration rates seem to cause a much slower spread of agents along the optimal path, resulting in a slowly damped ``wave-like" behaviour, as it can also be seen in the simulation examples in the supplementary multimedia file\footnote{This work has an attached multimedia file, available at http://ieeexplore.ieee.org, provided by the authors, showing the entire runs of the mean field system for different $\varepsilon$ parameters.}. These waves are caused by the initial conditions of agents, since all agents start at $\mathcal{S}$ on the first sub-graph, but they are more quickly damped (agents spreading out faster) for higher values of $\varepsilon$. This may have an impact when considering finite agent numbers; if we observe fast oscillations for a set of parameters as $n\to\infty$, there may be reasons to believe that these can result in non-convergent behaviour for finite agents.

We should also remark the interpretation of the mean field limit. By considering $\mathbf{y}(\infty):=\lim_{t\to\infty}\left(\lim_{n\to\infty} \hat{\mathbf{q}}(t,n)\right)$ we are computing the (limit) behaviour in time of an infinitely large system of agents. Our results do not guarantee, however, that the alternate limit $\lim_{n\to\infty}\left(\lim_{t\to\infty} \hat{\mathbf{q}}(t,n)\right)$ exists as well. The problem of studying this second limit corresponds to the limitations of a mean field approximation, and the study of stochastic trajectories of the finite agent system. Such study would shine some more light on how agents affect the limit distributions in these systems. It is worth mentioning that the impact of mean field solutions on discrete time MDPs is in itself a whole subject of study (see \cite{carmona2019model,carrillo2014derivation,gast2012mean}), and the interest on such mean field solutions applied to reinforcement learning problems seems to be growing fast in the last years. Knowing more about the relation between the distributions of finite agent systems and their mean field limits will give us tools to design multi-agent systems with guarantees concerning the number of agents needed to solve a specific problem.

\appendix
\subsection{Graph Doubling Procedure}\label{apx:graph}
Consider two identical graphs $\mathcal{G}^1,\,\mathcal{G}^2$ at $t=0$. Let $\hat{\mathbf{q}}_i^{1}(t)$ be the agent proportion in vertex $i\in \mathcal{G}^1$ with probability distribution $\hat{\mathbf{y}}^{1}(t)$, and $\hat{\mathbf{q}}_i^{2}(t)$ be the agent proportion in $\mathcal{G}^2$ such that $\hat{\mathbf{y}}^{2}(0)=\mathbf{0}$. Since the agents follow opposite weight fields (agents in graph 1 follow weights of graph 2, and viceversa), let us write the following matrices by considering the union of both systems, $\mathcal{G}^1\cup \mathcal{G}^2$:
\begin{equation*}\begin{aligned}
\mathbf{w}(t) := \left(\begin{array}{c}
\mathbf{w}^{1}(t)\\
\mathbf{w}^{2}(t)
\end{array}\right),
P^{\cup}(t,\varepsilon) := \left(\begin{array}{cc}
P^{2}(t,\varepsilon)& 0\\
0 &P^{1}(t,\varepsilon)
\end{array}\right),
\end{aligned}
\end{equation*}
and $\hat{\mathbf{y}}(t)\coloneqq \left(\begin{array}{l}
\hat{\mathbf{y}}^{1}(t)\\
\hat{\mathbf{y}}^{2}(t)
\end{array}\right).$ Note the reordering of the blocks in $P^{\cup}(t,\varepsilon)$, reflecting the fact that the agents follow opposite weights. The weight dynamics as written in \eqref{wdyn} are then
\begin{equation}\label{wdyn2}\begin{aligned}
\mathbf{w}_i^{1}(t+1)=(1-\rho)\mathbf{w}_i^{1}(t)+\rho \hat{\mathbf{q}}^{1}_i(t,n)R_i(t,r,\lambda),\\
\mathbf{w}_i^{2}(t+1)=(1-\rho)\mathbf{w}_i^{2}(t)+\rho \hat{\mathbf{q}}^{2}_i(t,n)R_i(t,r,\lambda),
\end{aligned}
\end{equation}
with $\mathbf{w}(0)=\mathbf{w}_0\mathbf{1}$. Agents enter graph $\mathcal{G}^2$ when they find the vertex $\mathcal{T}^1$, and go back to graph $\mathcal{G}^1$ when they find vertex $\mathcal{S}^2$, resulting in two interconnected systems having each an inflow ($\mathbf{u}^1(t),\mathbf{u}^2(t)\in\mathbb{P}^{|\mathcal{V}|}$) and outflow ($\mathbf{v}^1(t),\mathbf{v}^2(t)\in\mathbb{P}^{|\mathcal{V}|}$) of agents exiting and entering the graphs. The dynamics for the agent distribution can be written as
\begin{equation}\label{ydouble}\begin{aligned}
\hat{\mathbf{y}}^{1}(t+1)&=P^{2}(t,\varepsilon)\hat{\mathbf{y}}^{1}(t)+\mathbf{u}^{1}(t)+\mathbf{v}^{1}(t)\\
\hat{\mathbf{y}}^{2}(t+1)&=P^{1}(t,\varepsilon)\hat{\mathbf{y}}^{2}(t)+\mathbf{u}^{2}(t)+\mathbf{v}^{2}(t).
\end{aligned}
\end{equation}
Define now the selector matrices $S\in \mathbb{R}^{|V|\times|V|}$ and $T\in\mathbb{R}^{|V|\times|V|}$ as diagonal matrices with $S_{ii},T_{jj}=1$ for $i=\mathcal{S},j=\mathcal{T}$, zero otherwise. If $\mathbf{u}^1(t)$ is the distribution of agents entering graph $\mathcal{G}^1$ from graph $\mathcal{G}^2$ at time $t$ and $\mathbf{v}^1(t)$ is the density of agents leaving $\mathcal{G}^1$, both graphs are interconnected and closed to external inputs, and then:
\begin{equation}\label{inputoutput}\begin{aligned}
\mathbf{u}^{1}(t)&\equiv -\mathbf{v}^{2}(t)=SP^{1}(t,\varepsilon)\hat{\mathbf{y}}^{2}(t)\\
\mathbf{v}^{1}(t)&\equiv -\mathbf{u}^{2}(t)=TP^{2}(t,\varepsilon)\hat{\mathbf{y}}^{1}(t).
\end{aligned}
\end{equation}
Therefore, substituting \eqref{inputoutput} in \eqref{ydouble}, the agent probability distribution dynamics are given by
\begin{equation}\label{ygrouped}\begin{aligned}
\hat{\mathbf{y}}(t+1)=& \left(\begin{array}{cc}
(I-T)P^{2}(t,\varepsilon)&SP^{1}(t,\varepsilon)\\
TP^{2}(t,\varepsilon)&(I-S)P^{1}(t,\varepsilon)
\end{array}\right)\hat{\mathbf{y}}(t)=\\
=:&P(t,\varepsilon)\hat{\mathbf{y}}(t).
\end{aligned}
\end{equation}
Furthermore, observe that the matrix $P(t,\varepsilon)$ in \eqref{ygrouped} is also column stochastic. Effectively, we have interconnected the two graphs by the vertices $\mathcal{S}$ and $\mathcal{T}$, and made the agents move according to the opposite pheromones.
\subsection{Proofs}\label{apx:proofs}
\begin{proof}[Proof (Proposition \ref{prop:yinf})]
Given the bounded probability matrix $P(t,\varepsilon)$, for any edge $(ij)\in\mathcal{E}$, we have $P_{ji}(t,\varepsilon)\geq\varepsilon.$ Furthermore, since by Assumption \ref{as:1} there is at least one odd length cycle, the graph is aperiodic and we can directly invoke results from \cite{djarhscc} on convergence of stigmergy swarm probability distributions. In particular, from Theorem \ref{the:hscc}
\begin{equation*}
\exists \mathbf{y}(\infty) :\lim_{t\to\infty}\left(\prod_{t_k=0}^{t}P(t_k,\varepsilon) \right)\mathbf{y}(0) = \mathbf{y}(\infty) .
\end{equation*}
Since all positive terms in matrices $P(t,\varepsilon)$ are lower bounded, the product matrix $P^{\infty}(\varepsilon):=\lim_{t\to\infty}\prod_{t_k=0}^{t}P(t_k,\varepsilon)$ is irreducible, and from Theorem \ref{the:pft} the eigenvector $\mathbf{y}(\infty) $ is unique and has strictly positive entries.
Additionally, from Theorem \ref{the:hscc} we know that the convergence is exponential, with a rate bounded by $\alpha = (1-\frac{\varepsilon}{1+\left(g^{*}-1\right) \varepsilon}^{1+2\delta^*})^{\frac{1}{1+2\delta^*}}$.
\end{proof}
\begin{proof}[Proof (Lemma \ref{lem:1})]
First, since all $P(t,\varepsilon)$ have the positive entries lower bounded by $\varepsilon$ and the graph is connected, they are all irreducible and we can infer
\begin{equation}\label{eq:Pprodbound}\begin{aligned}
&\left(\prod_{t_k=t_0}^{t_0+2\delta^*}P(t_k,\varepsilon)\right)_{ji}=\left(P(t_0+2\delta^*,\varepsilon)...P(t_0,\varepsilon)\right)_{ji}\geq \varepsilon^{2\delta^*} \,\, \forall i,j\in\mathcal{V}.
\end{aligned}
\end{equation}
In other words, any vertex is reachable from any other vertex for times larger than $2\delta^*$. Now, making use of \eqref{eq:Pprodbound}, and $l^T_1, l^T_2,..., l^T_{|\mathcal{V}|}$ being the rows:
\begin{equation*}\begin{aligned}
&\prod_{t_k=t_0}^{t_0+2\delta^*}P(t_k,\varepsilon)  = \left(\begin{array}{c}l^T_1\\
...\\
l^T_{|\mathcal{V}|}
\end{array}\right)\Rightarrow \prod_{t_k=t_0}^{t_0+2\delta^*}P(t_k,\varepsilon)\mathbf{y}(t_0)= \left(\begin{array}{c}l^T_1 \mathbf{y}(t_0)\\
...\\
l^T_{|\mathcal{V}|} \mathbf{y}(t_0)
\end{array}\right)\geq \varepsilon^{2\delta^*}\mathbf{1}.
\end{aligned}
\end{equation*}
Therefore, for $t_0=0$ and $t>2\delta^*$ we have $\mathbf{y}(t)=\prod_{t_k=0}^{t}P(t_k,\varepsilon) \mathbf{y}(0) \geq \varepsilon^{t}\mathbf{1}.$ Last, from Proposition \ref{prop:yinf}, $\lim_{t\to\infty}\mathbf{y}(t)=\mathbf{y}(\infty) >\mathbf{0}$, therefore
\begin{equation*}
t>2\delta^*\Rightarrow \mathbf{y}(t)>\mathbf{0} \iff \operatorname{sgn}(\mathbf{y}(t))=\mathbf{1}.
\end{equation*}
\end{proof}

{\begin{proof}[Proof (Corollary \ref{prop:eig})]
From Theorem \ref{the:hscc} we know that the limit $\lim_{t\to\infty}\mathbf{y}(t+1)=\lim_{t\to\infty}\mathbf{y}(t)=\mathbf{y}(\infty)$ exists. Additionally, from Theorem \ref{the:1} we know that the limit $\lim_{t\to\infty}P(\varepsilon,t)=P(\varepsilon,\infty)$ also exists. Therefore, using the limit product rule:
\begin{equation*}\begin{aligned}
 \lim_{t\to\infty}\mathbf{y}(t+1) = \lim_{t\to\infty}  P (t, \varepsilon)\mathbf{y}(t)=P (\infty, \varepsilon)\mathbf{y}(\infty)=\mathbf{y}(\infty).
\end{aligned}
\end{equation*}
\end{proof}}
\begin{proof}[Proof (Proposition \ref{cor:1})]
From Theorem \ref{the:1}, the fixed point is $\mathbf{w}(\infty) = (I+\Gamma(r)+\lambda V(W(\infty)))\mathbf{1},$ and recall from Proposition \ref{prop:uniqueness} that it is unique. Additionally, $\mathbf{\gamma}_{\mathcal{S},\mathcal{T}}(r)=r$ and is $0$ for all other vertices, and it can be shown by contradiction (not added here for brevity) that $\operatorname{argmax}_i(\mathbf{w}_i(\infty) )= \mathcal{S},\mathcal{T}$. Now, to prove the proposition we assume the following structure for $\mathbf{w}(\infty)$, and later show it is indeed a solution (and therefore the only one, since it is unique). Let us assume  for $\mathbf{w}(\infty)$:
\begin{equation}\label{as:uniqueness}
v,u\in\mathcal{V}^1:\delta(\mathcal{S},v)>\delta(\mathcal{S},u)\Rightarrow \mathbf{w}_v(\infty)<\mathbf{w}_u(\infty),
\end{equation}
and the same holds for the converse $v,u\in\mathcal{V}^2$ with the distance to $\mathcal{T}$. That is, if $v$ is one step further away from $\mathcal{S}$ than $u$, then it has a smaller weight value. Now recall
\begin{equation}\label{eq:wopt2}
\mathbf{w}_i(\infty) = (1+\mathbf{\gamma}_i(r)+\lambda\max_{j\in\mathcal{V}}\mathbf{w}_{ij}(\infty)),
\end{equation}
and $\mathbf{\gamma}_i(r)=0\,\,\forall \,\,i\neq\mathcal{S},\mathcal{T}$. Then, $\forall j\in\mathcal{V}:\delta(\mathcal{S},j)=1$:
\begin{equation}\label{eq:wopt5}\begin{aligned}
\mathbf{w}_j(\infty) =& (1+\lambda\max_{k\in\mathcal{V}}\mathbf{w}_{jk}(\infty))= (1+\lambda \mathbf{w}_{\mathcal{S}}(\infty)),\\
\mathbf{w}_{\mathcal{S}}(\infty) =& (1+r+\lambda\max_{k\in\mathcal{V}}\mathbf{w}_{ik}(\infty))=(1+r+\lambda \mathbf{w}_j(\infty)).
\end{aligned}
\end{equation}
Solving \eqref{eq:wopt5} for both weights we obtain
\begin{equation}\label{eq:wopt6}\begin{aligned}
\mathbf{w}_{\mathcal{S}}(\infty) = \frac{1+r+\lambda}{1-\lambda^2},\,\,\mathbf{w}_j(\infty) = \frac{1+\lambda (1+r)}{1-\lambda^2}.
\end{aligned}
\end{equation}
Therefore, $r>0\Rightarrow \mathbf{w}_{\mathcal{S}}(\infty) >\mathbf{w}_j(\infty)\,\forall j:\delta(i,j)=1$. Then, for any $k\in\mathcal{V}^1,\, k\neq \mathcal{S}$,
\begin{equation}\label{eq:wopt7}\begin{aligned}
&\mathbf{w}_k(\infty) = 1+\lambda\max_{l\in\mathcal{V}}\mathbf{w}_{kl}(\infty)=1+\lambda +\lambda^2\max_{m\in\mathcal{V}}\mathbf{w}_{lm}(\infty)\\
=&...=\sum_{a=1}^{\delta(\mathcal{S},k)}\lambda^{a-1} + \lambda^{\delta(\mathcal{S},k)}\mathbf{w}_i(\infty) =\\
=& \sum_{a=1}^{\delta(\mathcal{S},k)}\lambda^{a-1} +\frac{\lambda^{\delta(\mathcal{S},k)}\left(1+r+\lambda\right)}{1-\lambda^2}=\frac{1+\lambda+ \lambda^{\delta(\mathcal{S},k)}r}{1-\lambda^2},
\end{aligned}
\end{equation}
and the same holds for any $k\in\mathcal{V}^2$ with the distance $\delta(\mathcal{T},k)$. Observe \eqref{eq:wopt7} yields an explicit solution to the fixed point $\mathbf{w}(\infty)$ that satisfies the assumption in \eqref{as:uniqueness}. From Proposition \ref{prop:uniqueness}, this is the only solution, thus \eqref{as:uniqueness} indeed holds for the fixed point and graphs considered. Finally, by construction \eqref{eq:wopt7} guarantees that picking the neighbouring maximum weight $\mathbf{w}(\infty)$ from any $v\in\mathcal{V}$ leads to $\mathcal{S}$ (or $\mathcal{T}$) through the minimum distance path, i.e. $\mathbf{w}(\infty)\in\mathcal{W}^*$.
\end{proof}
\begin{proof}[Proof (Proposition \ref{prop:optimal})]
From Definition \ref{def:boundedP}, if $\overline{\mathbf{y}}$ is the eigenvector of $P(\infty,0)$ corresponding to the eigenvalue $1$,
\begin{equation}\label{eq:eigenvecstar}\begin{aligned}
&P(\infty,0)\overline{\mathbf{y}} = \overline{\mathbf{y}}\Leftrightarrow \\
\Leftrightarrow &\left\{\begin{array}{c}
(I-T)P^{\nabla}(\mathbf{w}^2(\infty))\overline{\mathbf{y}}^1  + SP^{\nabla}(\mathbf{w}^1(\infty))\overline{\mathbf{y}}^2 =\overline{\mathbf{y}}^1 \\
TP^{\nabla}(\mathbf{w}^2(\infty)))\overline{\mathbf{y}}^1  + (I-S)P^{\nabla}(\mathbf{w}^1(\infty))\overline{\mathbf{y}}^2  = \overline{\mathbf{y}}^2 .
\end{array}\right.
\end{aligned}
\end{equation}
Recall Remark \ref{rem:disconnect}. Since we are considering the full doubled graph with $|\mathcal{V}|=2|\mathcal{V}_1|=2|\mathcal{V}_2|$ (that is, with all $\mathcal{T}_1,\mathcal{S}_1,\mathcal{T}_2,\mathcal{S}_2\in\mathcal{V}$), there are two vertices in the graph that are effectively disconnected from the rest, namely $\mathcal{T}_1$ and $\mathcal{S}_2$. Therefore, $y_{\mathcal{T}_1}(t)=y_{\mathcal{S}_2}(t)=0\quad \forall\,t.$ Similarly, 
\begin{equation}\label{eq:ystar12}
\mathcal{T}_1,\mathcal{S}_2\notin \left(\cup p^*_{ST}\right)\cup\left(\cup p^*_{TS}\right)\Rightarrow \overline{\mathbf{y}}_{\mathcal{T}_1}=\overline{\mathbf{y}}_{\mathcal{S}_2} = 0.
\end{equation}
Let us focus on the first equality in \eqref{eq:eigenvecstar}. Recall $\overline{\mathbf{y}}^1_{\mathcal{S}}=\overline{\mathbf{y}}^2_{\mathcal{T}}=\frac{1}{2k}$ and $\overline{\mathbf{y}}^1_{\mathcal{T}}=\overline{\mathbf{y}}^2_{\mathcal{S}}=0$. Let us now verify that $P^{\nabla}(\mathbf{w}^1(\infty))\overline{\mathbf{y}}^2=\overline{\mathbf{y}}^2$ for all vertices $v\neq \mathcal{S},\mathcal{T}$. Recall from Definition \ref{def:pnabla} that $P^{\nabla}_{ji}(\mathbf{w}^1(\infty))=\frac{1}{|N_i^{out}|}$  $\forall \, j\in N_i^{out}$. Then, for any $v\neq \mathcal{S},\mathcal{T}$,
\begin{equation}\label{eq:prodnabla1}\begin{aligned}
&\left(P^{\nabla}(\mathbf{w}^1(\infty))\overline{\mathbf{y}}^2\right)_v = \sum_{j\in N_v^{in}}\frac{\overline{\mathbf{y}}_j^2}{|N_j^{out}|}.
\end{aligned}
\end{equation}
Substituting now $\overline{\mathbf{y}}_j^2 = \frac{1}{2k}\sum_{p\in\pi_{Tj}}\prod_{u\in p\setminus j}\frac{1}{|N_u^{out}|} $ in \eqref{eq:prodnabla1}:
\begin{equation}\label{eq:prodnabla2}\begin{aligned}
\sum_{j\in N_v^{in}}\frac{\overline{\mathbf{y}}_j^2}{|N_j^{out}|}=&\sum_{j\in N_v^{in}}\frac{1}{2k}\left(\sum_{p\in\pi_{Tj}}\prod_{u\in p\setminus j}\frac{1}{|N_u^{out}|}\right)\frac{1}{|N_j^{out}|}=\\
=&\frac{1}{2k}\sum_{j\in N_v^{in}}\sum_{p\in\pi_{Tj}}\prod_{u\in p}\frac{1}{|N_u^{out}|}.
\end{aligned}
\end{equation}
Since all $j\in N_v^{in}$ lead to $v$, \eqref{eq:prodnabla2} is simply
\begin{equation}\begin{aligned}
\frac{1}{2k}\sum_{j\in N_v^{in}}\sum_{p\in\pi_{Tj}}\prod_{u\in p}\frac{1}{|N_u^{out}|} = \frac{1}{2k}\sum_{p\in\pi_{Tv}}\prod_{u\in p\setminus v}\frac{1}{|N_u^{out}|}=\overline{\mathbf{y}}^2_v, 
\end{aligned}
\end{equation}
and $\left(P^{\nabla}(\mathbf{w}^1(\infty))\overline{\mathbf{y}}^2\right)_v= \overline{\mathbf{y}}^2_v$. Similarly, 
\begin{equation}\label{eq:eigstar}
\left(P^{\nabla}(\mathbf{w}^2(\infty))\overline{\mathbf{y}}^1\right)_v = \overline{\mathbf{y}}^1_v\quad \forall v\neq \mathcal{S},\mathcal{T},
\end{equation}
and $\overline{\mathbf{y}}^1_\mathcal{T}=0$. Now observe
\begin{equation}\label{eq:eigstar1}
\left(SP^{\nabla}(\mathbf{w}^1(\infty))\overline{\mathbf{y}}^2 \right)_i=\left\{\begin{array}{l}
\left(P^{\nabla}(\mathbf{w}^1(\infty))\overline{\mathbf{y}}^2\right)_i\quad \text{if}\,i=\mathcal{S},\\
0\quad\text{else}.
\end{array}\right.
\end{equation}
From Proposition \ref{cor:1}, we know that $\mathbf{w}_{\mathcal{S}}^1 (\infty) = \max_{j}\mathbf{w}_j (\infty)\Rightarrow P^{\nabla}_{\mathcal{S}i}(\mathbf{w}^1(\infty)) =1 \,\,\forall \,\,(i\mathcal{S})\in\mathcal{E}$. Since all paths $p\in\pi_{TS}$ start and end at the same vertices and have the same length, recall $\frac{1}{|N_u^{out}|}$ can be interpreted as the probability of moving out of $u$, therefore the product $\Pr\{p\}:=\prod_{u\in p\setminus \mathcal{S}}\frac{1}{|N_u^{out}|}$ is the probability of following the entire path $p$, and it holds that 
\begin{equation}\label{eq:eigstar2}\begin{aligned}
\sum_{p\in\pi_{TS}}\prod_{u\in p\setminus \mathcal{S}}\frac{1}{|N_u^{out}|}=\sum_{p\in\pi_{TS}}\Pr\{p\}=1,
\end{aligned}
\end{equation}
Therefore, by making use of \eqref{eq:eigstar2}, we can compute \eqref{eq:eigstar1}:
\begin{equation}\label{eq:eigstar3}\begin{aligned}
&\left(SP^{\nabla}(\mathbf{w}^1(\infty))\overline{\mathbf{y}}^2\right)_\mathcal{S} =\sum_{j\in N_\mathcal{S}^{in}}\frac{\overline{\mathbf{y}}_j^2}{|N_j^{out}|}=\sum_{j\in N_\mathcal{S}^{in}}\frac{1}{2k}\left(\sum_{p\in\pi_{Tj}}\prod_{u\in p\setminus j}\frac{1}{|N_u^{out}|}\right)\frac{1}{|N_j^{out}|}=\\
=&\frac{1}{2k}\sum_{j\in N_\mathcal{S}^{in}}\sum_{p\in\pi_{Tj}}\prod_{u\in p}\frac{1}{|N_u^{out}|}=\frac{1}{2k}\sum_{p\in\pi_{TS}}\prod_{u\in p\setminus \mathcal{S}}\frac{1}{|N_u^{out}|}=\frac{1}{2k}=\overline{\mathbf{y}}^1_\mathcal{S}.
\end{aligned}
\end{equation}
At last, combining \eqref{eq:eigstar} and \eqref{eq:eigstar3} we have 
\begin{equation}
(I-T)P^{\nabla}(\mathbf{w}^2(\infty))\overline{\mathbf{y}}^1  + SP^{\nabla}(\mathbf{w}^1(\infty))\overline{\mathbf{y}}^2 =\overline{\mathbf{y}}^1 ,
\end{equation}
and analogously one can show that the same holds for the second equation in \eqref{eq:eigenvecstar}. Therefore, $P(\infty,0)\overline{\mathbf{y}} = \overline{\mathbf{y}}$.
\end{proof}

\section*{ACKNOWLEDGMENTS}
Authors would like to thank G. Delimpaltadakis, G. Gleizer and C. Verdier for the useful discussions. This work was supported by the ERC Starting Grant SENTIENT 755953.
\bibliographystyle{ACM-Reference-Format}
\bibliography{MF_Robotics.bib}

\end{document}